%% file: main.tex
\documentclass[10pt,a4paper]{article}
\pdftrailerid{}
\usepackage[T1]{fontenc}
\usepackage[utf8]{inputenc}
\usepackage[dvipsnames]{xcolor}
\usepackage{soul}
\usepackage{amsmath,amssymb,graphicx,booktabs,tabularx,array}
\usepackage{caption,float,placeins,needspace,enumitem}
\usepackage[most]{tcolorbox}
\usepackage[numbers,sort&compress]{natbib}
\usepackage{style/piccolo-ty}
\usepackage{bookmark}
\usepackage{footnotehyper}

\usepackage{todonotes}

\makesavenoteenv{tcolorbox}
\definecolor{picink}{HTML}{293C55}
\definecolor{picblue}{HTML}{607DA2}
\definecolor{piclight}{HTML}{F1F4F8}
\definecolor{sectioncolor}{HTML}{505D90}
\definecolor{subsectioncolor}{HTML}{606FA0}
\definecolor{mypurple}{HTML}{505D90}
\definecolor{absborder}{HTML}{8996BA}
\definecolor{absbg}{HTML}{F5F6FB}

\colorlet{takeawaybg}{picblue!11!white}
\colorlet{takeawaylabelbg}{picblue!27!white}
\DeclareRobustCommand{\takeaway}[1]{%
  \unskip\space\begingroup\upshape
  \textcolor{picink}{\sethlcolor{takeawaylabelbg}\hl{\textbf{Takeaway:}}}~%
  \sethlcolor{takeawaybg}\hl{#1}%
  \endgroup
}

\setcitestyle{numbers,square,citesep={,}}
\hypersetup{pdftitle={Piccolo: Co-Evolving Decision Models and Data through Recursive Self-Improvement
},pdfauthor={Piccolo Team},colorlinks=true,linkcolor=picblue,citecolor=picblue,urlcolor=picblue}
\renewcommand{\arraystretch}{1.15}
\newcolumntype{Y}{>{\raggedright\arraybackslash}X}

\setlist[itemize]{leftmargin=1.2em,itemsep=2pt,topsep=3pt,parsep=0pt}
\makeatletter
\let\picsection\section
\renewcommand{\section}{\Needspace{6\baselineskip}\picsection}
\let\picsubsection\subsection
\renewcommand{\subsection}{\Needspace{4\baselineskip}\picsubsection}
\makeatother
\title{{\fontsize{19}{23}\fontfamily{phv}\selectfont\bfseries\color{black}Piccolo: Co-Evolving Decision Models and Data \\ through Recursive Self-Improvement}}
\DeclareRobustCommand{\teamlink}{\textsuperscript{\hyperlink{piccolo-team}{\textdagger}}}
\author{\normalsize Piccolo Team\protect\teamlink}
\date{}
\begin{document}
\maketitle
\vspace{-27pt}
\begingroup\centering
\small\sffamily
\href{https://huggingface.co/collections/agentscope-ai/piccolo}{\raisebox{-2pt}{\includegraphics[height=11pt]{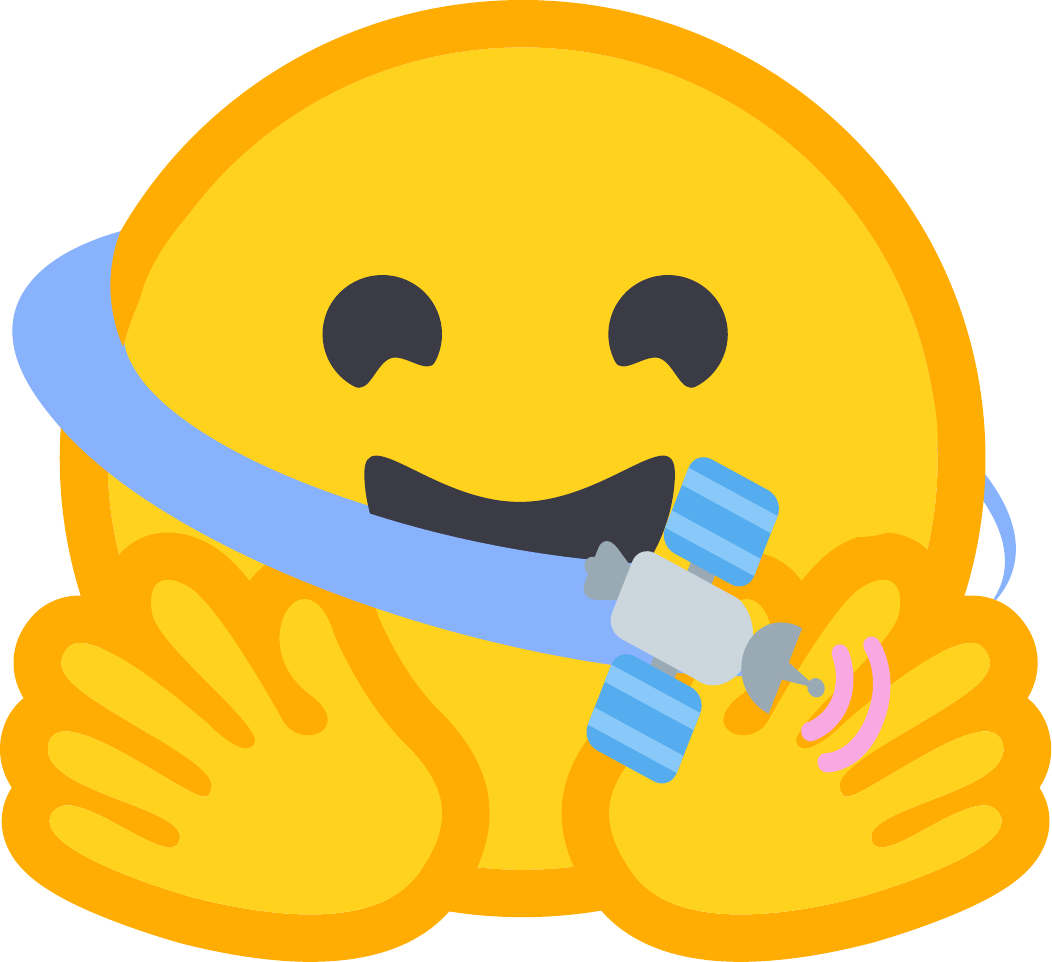}}\;Models}
\hspace{1.8em}
\href{https://github.com/agentscope-ai/Piccolo}{\raisebox{-2pt}{\includegraphics[height=10pt]{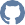}}\;Code}
\hspace{1.8em}
\href{https://decisionplayground.agentscope.io/piccolo.html}{\raisebox{-2pt}{\includegraphics[height=10pt]{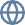}}\;Blog}
\par\endgroup
\vspace{-3pt}
\begin{abstract}
\input{sections/00_abstract}

\end{abstract}
\input{sections/01_introduction}
\input{sections/02_data}
\input{sections/03_rsi}

\input{sections/04_evolution}

\input{sections/05_discussion}

\input{sections/06_outlook}

\label{maintext:end}

\begingroup
\normalsize
\input{sections/team}

\endgroup

\clearpage

\clearpage
\begingroup
\renewcommand{\bibfont}{\small}
\setlength{\bibsep}{3pt}
\bibliographystyle{unsrtnat}
\bibliography{references}
\endgroup
\clearpage
\appendix
\fontsize{10}{12.5}\selectfont
\input{sections/appendix/full_public_rankings}
\input{sections/appendix/candidate_supervision}
\input{sections/appendix/data_recipes}
\input{sections/appendix/evaluation_details}
\input{sections/appendix/research_records_compute}
\end{document}

%% file: sections/00_abstract.tex
We present Piccolo, a family of multimodal decision models (0.8B to 27B) that co-evolves model capabilities and training data via Agentic Data Engineering and recursive self-improvement (RSI).
The models predict probabilities over candidate answers for general-purpose text and visual judgments, agent harnesses, data workflows, and interactive games. Piccolo-27B tops the Decision Index Vision Full leaderboard with a score of 75.59, while Piccolo-9B and Piccolo-4B achieve 68.35 and 67.46, respectively. All three rank first on the Private benchmark in their respective model-size categories.\footnote{\textbf{Decision-Index-Vision snapshot: 2026-11-17 18:35:34 UTC}, with 86 entrants. Full-score is the equal-weight mean of Public and Private. Size-specific ranks use nominal total-parameter classes \cite{decisionboard}.} Research agents analyze model behavior, construct training data, conduct experiments, and use the results to refine subsequent proposals. Semantic slices and a shared research memory connect these actions across the model family's development. With training recipes comprising hundreds of thousands of examples, data selection and feedback-driven refinement are central to the self-improvement process. 
Our large-scale RSI exploration yields three counter-intuitive insights for decision-model development: (1) Backbone-recipe co-adaptation outweighs raw model recency --- earlier backbones can surpass newer generations on visual tasks under identical recipes; (2) Targeted data curation enables 4B–9B models to rival 27B baselines on specialized workflows; and (3) Continuous loss reduction can decouple from and even degrade discrete decision accuracy.
Together with the released weights and tools from the Data-Juicer and AgentScope ecosystems, Piccolo provides an actionable blueprint for building frontier decision models and scaling data-driven RSI.

%% file: sections/01_introduction.tex
\begin{figure}[H]
\centering
\includegraphics[width=\linewidth]{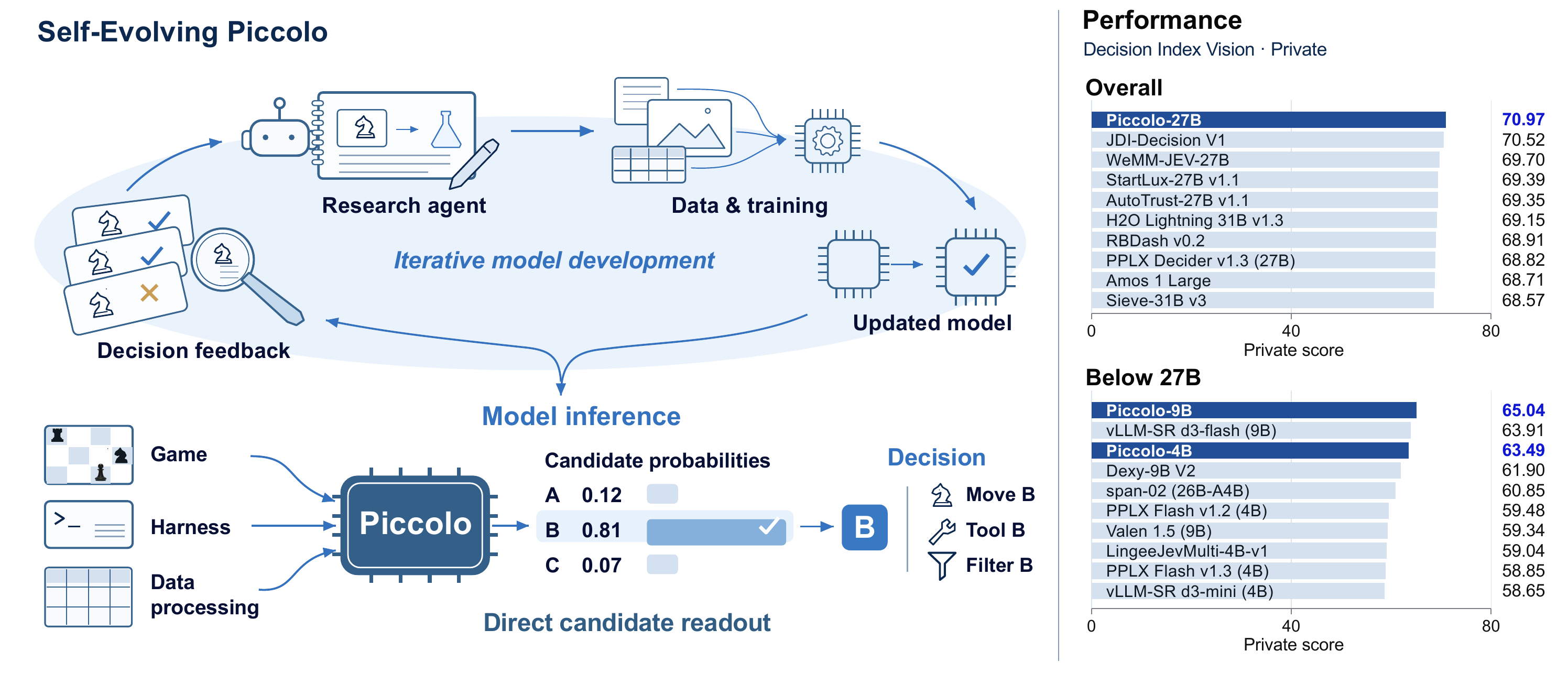}
\caption{Piccolo overview and decision performance. Left: agentic self-improvement through iterative feedback, research, data and training, alongside direct candidate-probability readout for  various decisions. Right: Decision Index Vision \textbf{\emph{Private set scores}}, showing the top ten overall (top) and below 27B (bottom). Full/Public comparisons are reported in Appendix~\ref{app:full_public_rankings}.}
\label{fig:public_vision}
\end{figure}

\section{Introduction}

\begin{figure}[!tbp]
\centering
\includegraphics[width=\linewidth]{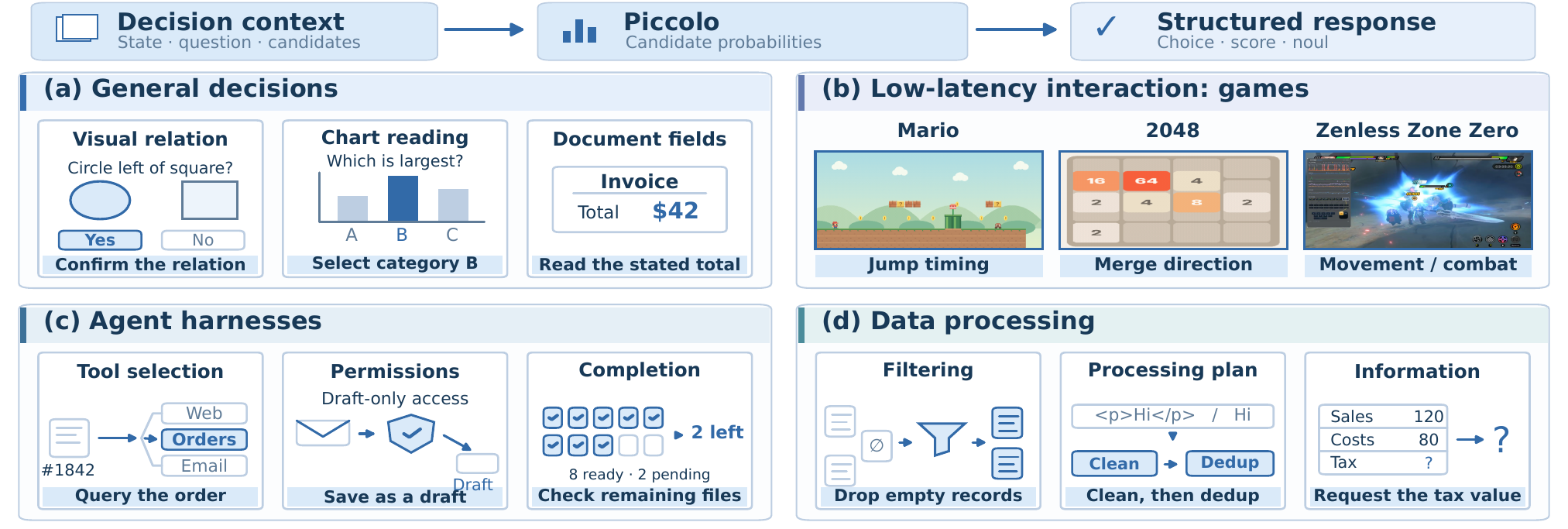}
\caption{Piccolo applications: enterprise decisions, embodied interaction through game examples, agent harnesses, and data processing. Appendix~\ref{app:evaluation_details} expands the workflow examples.}
\label{fig:capabilities}
\end{figure}

While generative models excel at open-ended synthesis, real-world autonomous systems bottleneck on decisions, such as selecting actions, enforcing guardrails, and routing workflows under strict latency and reliability constraints. 
A \textbf{decision model} maps a supplied state, question, and candidate set to probabilities over answers or actions \cite{mammoth,drex,startlux}.
We present Piccolo, which implements this Jev-like interface for text and images through direct candidate-logit readout from a language-model head. Its System-1-style predictions support \textbf{Choice} over alternatives, \textbf{Score} on ordered levels, and \textbf{Noul} for yes/no judgments. Callers can select actions, threshold confidence, or escalate decisions using the returned distribution; Appendix~\ref{app:candidate_supervision} gives the formulation.

% \textbf{Piccolo spans 0.8B, 2B, 4B, 9B, and 27B.} The 27B, 9B, and 4B models score \textbf{70.97}, \textbf{65.04}, and \textbf{63.49} on Private, leading their parameter classes (Figure~\ref{fig:public_vision}). Their Full scores are \textbf{75.59}, \textbf{68.35}, and \textbf{67.46}: first overall for 27B, second and third below 27B for 9B and 4B. 

Piccolo powers four core application domains (Figure~\ref{fig:capabilities}): \textbf{(a) General decisions:} visual interpretation, document checks, and routing for enterprise workflows with locally retained sensitive data. \textbf{(b) Low-latency interaction like games:} repeated perception--action decisions support movement and combat in Zenless Zone Zero, jump timing in Mario, and merge choices in 2048 \cite{piccoloweb}. \textbf{(c) Agent harnesses:} tool review, permissions, and context management in systems built with AgentScope \cite{agentscope}; QwenPaw integrates Piccolo for tool review and context compaction \cite{qwenpawpiccolo}. \textbf{(d) Data processing:} filtering, operator selection, and quality judgments within Data-Juicer workflows \cite{datajuicer}. HarnessJevBench and DataJevBench organize recurring harness and data decisions into complementary evaluation suites \cite{harnessbench,databench}.

Spanning five sizes from 0.8B to 27B, Piccolo establishes a new Pareto frontier on Decision Index Vision (Figure~\ref{fig:pareto}). Piccolo-27B ranks \#1 overall (75.59 Full, 70.97 Private); 9B and 4B lead their Private parameter classes (65.04 and 63.49). The 0.8B and 2B models score 55.17 and 66.64 Public, respectively. Piccolo-27B also ranks among the leading models on HarnessJevBench~\cite{harnessbench} (84.41) and DataJevBench~\cite{databench} (82.60).

% \bolinnotes{In the above paragraph, make sure the numbering ``(1) v.s. (a)'' and the captions ``Enterprise decisions v.s. General decisions'' are consistent with those in Figure 2.}

\textbf{Agentic Data Engineering drives Piccolo's RSI.} Training dynamics and fine-grained data buckets characterize where the model learns reliably, where progress stalls, and which capabilities compete. Linking these responses to domain, supervision, and task structure turns an aggregate score into a map of capability boundaries. Agents use this map to propose data changes, build experiments, and carry the resulting knowledge into the next research decision \cite{ade}. Researchers interactively guide objectives and introduce methods while agents automatically revise mixtures, synthesis targets, and training choices. Shared assets preserve discoveries and lessons throughout evolving progress. 
%The 27B model development involved training experiments equivalent to approximately \(2.8 \times 10^{21}\) FLOP at the hardware's peak BF16 performance.\footnote{Estimated from approximately 6,300 PPU-ZW810E accelerator-hours using a reference BF16 peak rate of 123 TFLOP/s per device. Hardware references, accounting assumptions are detailed in Appendix~\ref{app:compute_accounting}.}

% Several findings motivate the analysis that follows. \textbf{Data selection changes the capability profile}: finer selection improved workflow decisions while a broader adaptive mixture favored Text (Section~\ref{sec:data-proposals}). \textbf{A recipe and its model must be selected together}: the same data supported different strengths across backbones (Figure~\ref{fig:recipe_gains}). \textbf{Research value extends beyond the highest-scoring branch}: task regressions, computation costs, and reusable update directions inform subsequent proposals (Section~\ref{sec:discussion}). These observations connect the RSI design to the family-level results and the experiments behind them.

Across $\sim$600 experimental runs (approximately \(2.8 \times 10^{21}\) FLOPs at the hardware's peak BF16 performance)\footnote{Estimated from approximately 6,300 PPU-ZW810E accelerator-hours using a reference BF16 peak rate of 123 TFLOP/s per device. Hardware references and accounting assumptions are detailed in Appendix~\ref{app:compute_accounting}.}, our agent-led loop uncovered development dynamics that challenge standard post-training heuristics:
(1) \textbf{Data granularity dictates capability trade-offs}: Fine-grained filtering sharpens agentic workflow decisions, whereas broader adaptive mixtures favor general text reasoning (Section~\ref{sec:data-proposals}).
(2) \textbf{Recipes and backbones must be co-selected}: Identical data mixtures unlock divergent strengths across different model generations (Figure~\ref{fig:recipe_gains}).
(3) \textbf{Failed branches hold latent value}: Even regressive runs yield reusable parameter update directions and diagnostic signals that accelerate subsequent iterations (Section~\ref{sec:discussion}).
These observations connect the RSI design to the family-level results and the experiments behind them. In summary, our main contributions are threefold: 
\begin{itemize}
\item \textbf{A multimodal decision model family:} competitive size–quality trade-offs, leading Vision Full and size-specific Private results, and applications from enterprise workflows to interactive control.
\item \textbf{A data-driven RSI methodology:} fine-grained feedback, explicit proposals, and reusable experiments turn model development into a continuing agent-led research loop.
\item \textbf{Empirical insights for decision model development:} systematic findings connect supervision, validation coverage, synthesis, backbone selection, and post-training to concrete experimental outcomes.
\end{itemize}

%% file: sections/02_data.tex
\FloatBarrier
\section{Preliminaries}

\begin{figure}[!htbp]
\centering
\includegraphics[width=\linewidth]{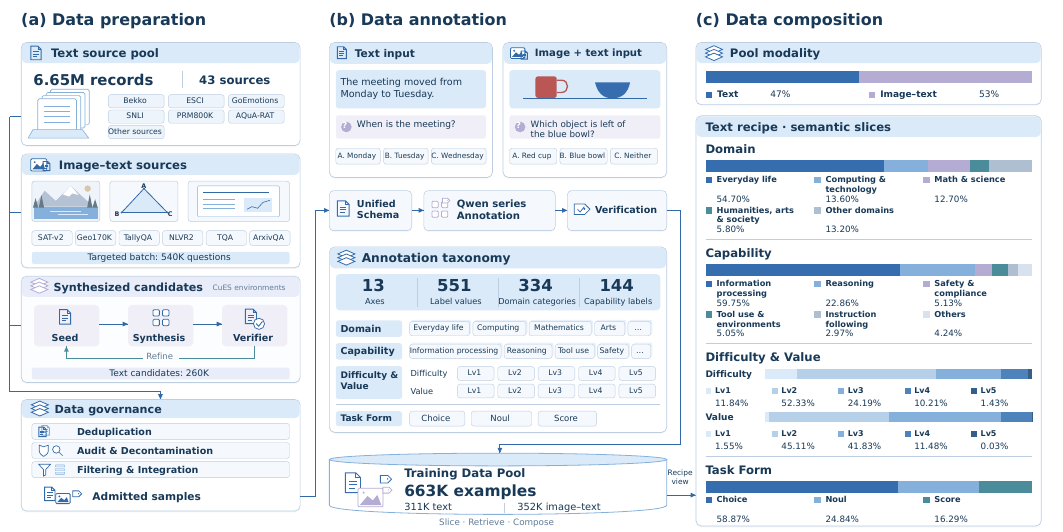}
\caption{Data preparation, semantic annotation, and composition. Collected and synthesized material passes through common governance into a versioned training pool. The modality split covers the joint text and image--text pool; label distributions describe a representative text recipe.}
\label{fig:data}
\end{figure}

\subsection{Decision Model}
\label{sec:definition}

The decision model maps a state, a question, and an ordered candidate set to a probability distribution. A training example is $(s,u,C,q)$, where $s$ contains text, images, or structured observations, $u$ specifies the question, $C=(c_1,\ldots,c_K)$ lists the candidates, and $q$ is their target distribution. The three interfaces share this representation: \textbf{Choice} compares supplied alternatives, \textbf{Noul} uses a yes/no pair, and \textbf{Score} uses ordered rating levels.

Piccolo combines a visual encoder, a connector, and a language backbone with its native language-model head. The backbone uses bidirectional attention over the prompt in full-attention blocks. In hybrid backbones, linear-attention blocks retain their original causal computation. To construct the input $x=\operatorname{compile}(s,u,C)$, each candidate $c_k$ is assigned a single-token label $\tau(k)$ and presented with its description. Images are encoded by the visual encoder and passed through the connector to the language backbone alongside the text input. An optional sidecar OCR model supplies extracted text as an additional image-associated observation.

At the answer position, the backbone produces a hidden state $h_\theta(x)$. The native language-model head projects this state to vocabulary logits through $W_{\mathrm{LM}}$. Piccolo reads the logit for each candidate label $\tau(k)$ and applies a softmax over these $K$ logits to obtain the decision distribution:
\begin{equation}
\label{eq:candidate_readout}
z_{\theta,k}(x)=\bigl[W_{\mathrm{LM}}h_\theta(x)\bigr]_{\tau(k)},\qquad
p_{\theta,k}(x)=\frac{\exp z_{\theta,k}(x)}{\sum_{j=1}^{K}\exp z_{\theta,j}(x)}.
\end{equation}
Appendix~\ref{app:candidate_supervision} details input organization and the supervision--error relationship.

\subsection{Training and Optimization}
\label{sec:training}

Supervised training minimizes cross-entropy over the candidate distribution for both hard and soft targets:
\begin{equation}
\label{eq:candidate_supervision}
\ell_{\mathrm{CE}}(q,p_\theta)=-\sum_{k=1}^{K}q_k\log p_{\theta,k},\qquad
\ell_{\mathrm{CE}}-H(q)=D_{\mathrm{KL}}(q\Vert p_\theta).
\end{equation}
For a one-hot target, the loss is the negative log-probability of the correct answer. For soft targets, subtracting target entropy separates distributional mismatch from uncertainty in the supervision. Cross-entropy is a proper scoring rule, providing a basis for calibration-oriented supervision~\cite{multiclasslogloss}. Accuracy additionally records whether the correct candidate ranks first. Section~\ref{sec:rsi} uses these complementary signals to guide data changes.

We use candidate-level reinforcement learning to refine decision accuracy in models after supervised training. These continuations optimize the candidate distribution with correctness-based, group-relative updates and supervised replay. % Appendix~\ref{app:data_recipes} summarizes the refinement objective.

We use weight-space optimization to search for improved decision accuracy by directly modifying the trained model parameters. Candidate weights are constructed by interpolating checkpoints, rescaling parameter projections, or applying directions extracted from previous training updates~\cite{evolutionarymerge}. This allows us to explore combinations and adjustments of learned parameters without running a new gradient-based training stage for each candidate. Appendix~\ref{app:data_recipes} summarizes more details of RL objective and these refinement methods.

\subsection{Data Sources and Preparation}
\label{sec:data}

To support generalization across domains, task structures, and input modalities, we collect a broad range of text and visual decision data. The collection spans everyday decisions, computing and technology, mathematics and science, humanities, and arts. Existing sources and synthesized tasks cover reasoning, retrieval, preference judgments, visual interpretation, and action selection. Appendix~\ref{app:source_inventory} provides the source inventory.

Training on these heterogeneous sources requires preserving the evidence and conditions that justify each target while making their supervision consistent with the shared decision interface. Conversion places each question in the $(s,u,C,q)$ format, retaining action preconditions, relevance distinctions, image--answer alignment, and the original supervision distribution. Data-Juicer supports normalization, filtering, and auditing~\cite{datajuicer}. We remove duplicates to limit redundant training exposure. Task-specific audits repair or exclude incomplete and incorrect  data that would otherwise introduce misleading supervision. To extend coverage, we use CuES for data synthesis~\cite{cues}.

The resulting pool shown in Figure~\ref{fig:data} contains 663K examples, comprising 311K text and 352K image--text decisions. Appendix~\ref{app:data_recipes} details the data composition. We annotate examples by domain, capability, task function, difficulty, and learning value. These labels group examples into semantic slices whose training dynamics and decision accuracy can be analyzed together. Each example retains its identity and source, allowing agents to trace slice-level findings back to the underlying data for inspection, revision, or targeted synthesis. Section~\ref{sec:rsi} describes this feedback loop, and Appendix~\ref{app:semantic_taxonomy} provides the annotation scheme.

%% file: sections/03_rsi.tex
\clearpage
\section{Recursive Self-Improvement with Data-driven Agents}
\label{sec:rsi}

In conventional model development, failed training runs are discarded as sunk compute costs. In Piccolo’s RSI process, we capture and organize every successful or regressive experiment as structured epistemic capital. 
By binding fine-grained semantic slices, training dynamics, and weight-space trajectories into a persistent research memory, our agent orchestrator transforms brute-force hyperparameter search into cumulative scientific discovery.
% We organize Piccolo's recursive self-improvement around proposals that translate feedback into data and model experiments. The orchestrator in Figure~\ref{fig:rsi} connects this loop to structured feedback, human guidance, shared assets, and executable experiments. This section defines the research objective and atomic loop, describes the surrounding design, and explains their use in Piccolo development.

\subsection{Evolving Objective and Iteration Loop}
\label{sec:rsi-objective}

\paragraph{Evolving objective.}
We seek to improve Piccolo's decision performance through experiments on data strategies and model configurations. Let $S_{\mathrm{dev}}(m;\mathcal V)$ measure performance on the candidate-based tasks defined in Section~\ref{sec:definition}, using a held-out development set $\mathcal V$ for each research phase. The objective is
\begin{equation}
\label{eq:rsi-objective}
\max_{\psi_{1:T}}\;
S_{\mathrm{dev}}\bigl(m_T^{\mathrm{best}};\mathcal V\bigr)
\qquad\text{subject to}\qquad
\sum_{t=1}^{T}c_t\leq B,
\end{equation}
where $\psi_t$ is an experimental proposal and $m_T^{\mathrm{best}}$ is the best evaluated model, including the starting model, after $T$ experiments. The number of allocated accelerators and the exploration window define $B$ in accelerator-hours; $c_t$ is the execution and evaluation cost of experiment $t$ in the same units. Each proposal specifies a starting model, a data strategy, and permitted training or inference changes. Data strategies determine source selection, mixtures, revisions, and synthesis. Models are compared under a common scoring rule, with independent confirmation assessing transfer (Section~\ref{sec:rsi-heldout}).

\paragraph{Atomic iteration loop.}
Figure~\ref{fig:rsi} organizes the cycle as \emph{proposal, build and experiment, results, and diagnosis}. Let $\mathcal R_t$ record the available models, annotated data assets, completed comparisons, research memory, and remaining budget. Agents use this state to turn findings into subsequent experiments~\cite{ade}:
\begin{equation}
\label{eq:rsi-loop}
\mathcal R_{t-1}
\xrightarrow{\mathrm{Propose}}\psi_t
\xrightarrow{\mathrm{Experiment}}(m_t,\mathbf f_t,c_t)
\xrightarrow{\text{Diagnose \& Update}}\mathcal R_t.
\end{equation}
Feedback $\mathbf f_t$ summarizes learning dynamics, inference results, and data statistics by semantic slice. Diagnosis connects gains and regressions to these slices and the tested settings. The update records resulting assets and findings in shared research memory and revises proposal priorities. For example, a mastered synthetic slice can prompt harder variants. Human researchers can optionally introduce new directions, set exploration priorities and budgets. Subsequent proposals can start from any retained model and reuse findings across historical branches.

\subsection{Orchestrator Design around the Loop}
\label{sec:rsi-design}

\begingroup
\setlength{\parskip}{4pt}
\setlength{\intextsep}{6pt plus 1pt}
Running the atomic loop requires coordinating feedback, data and model assets, execution, and human guidance. We organize these functions into an orchestrator, maintaining $\mathcal R_t$ through versioned records of experiment lineage and proposal priorities across concurrent branches (Figure~\ref{fig:rsi}).

\begin{figure}[!tbp]
\centering
\includegraphics[width=\linewidth]{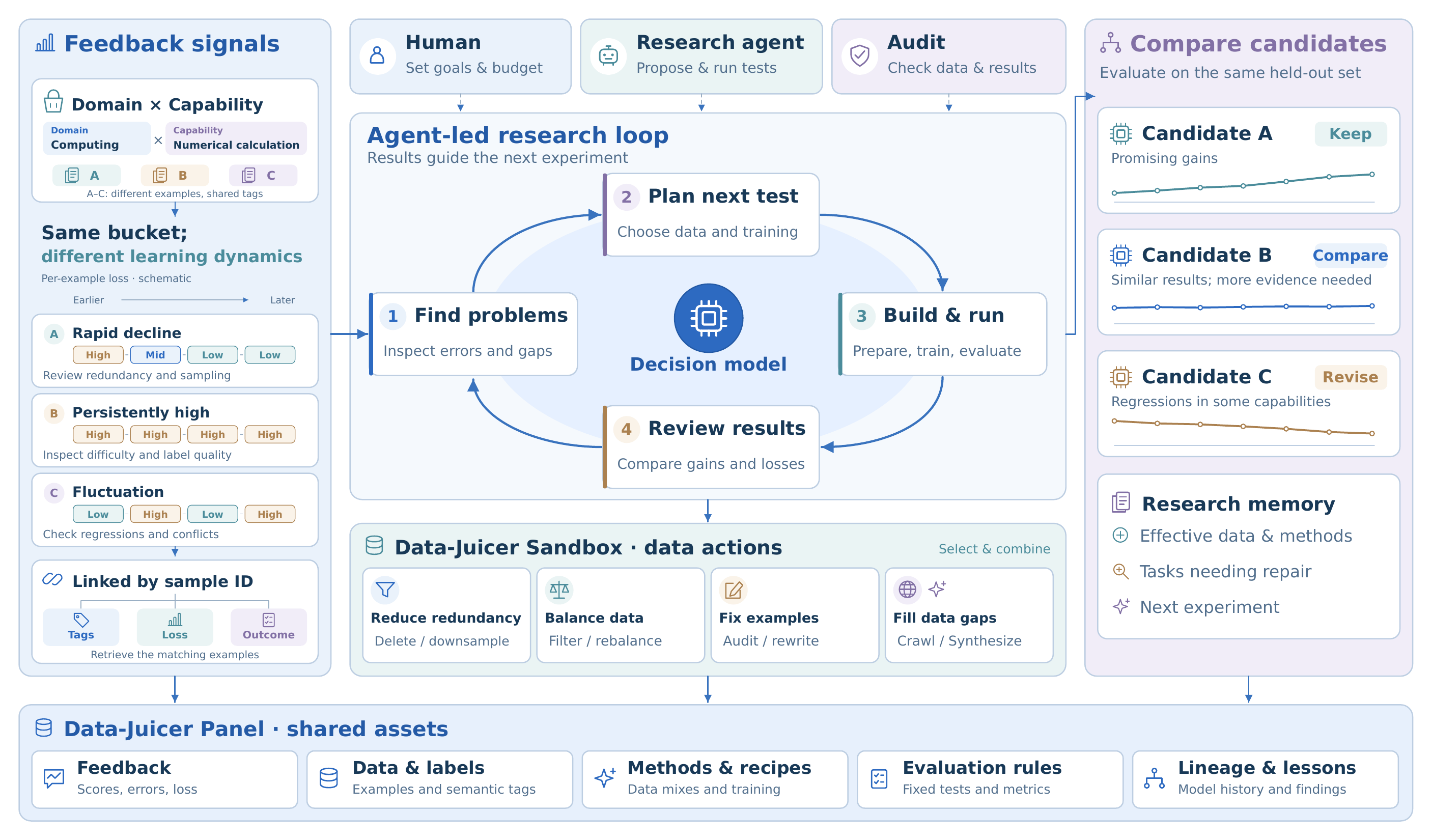}
\caption{The orchestrator of Piccolo's RSI links feedback (left), an experiment loop (center), shared assets (bottom), and comparison with memory (right). Agents use feedback and accumulated findings to plan subsequent experiments; human researchers set goals and review major changes.}
\label{fig:rsi}
\end{figure}

\paragraph{Structured Feedback.}
Semantic labels align training dynamics, inference outcomes, and data statistics by source, supervision, and candidate count. Hard-target cross-entropy tracks the probability assigned to the correct candidate~\cite{multiclasscalibration,multiclasslogloss}; for soft targets, subtracting target entropy yields target--prediction KL (Appendix~\ref{app:candidate_supervision}). Loss trajectories describe how the predicted distribution evolves, while repeated inference tracks repaired and newly missed decisions. Aggregating these signals by semantic slice reveals differences in learning progress among examples with shared domain and capability tags. The agent links these observations to sample records to identify coverage gaps, uneven learning, and label conflicts. These findings guide the data actions in Figure~\ref{fig:rsi}: downsampling redundant examples, rebalancing mixtures, revising conflicting labels, and synthesizing examples for underrepresented capabilities.

\begin{table}[H]
\centering
\newsavebox{\workflowfeedbacktextbox}
\begin{lrbox}{\workflowfeedbacktextbox}
\begin{minipage}[t]{0.44\linewidth}
\vspace{0pt}
\normalsize
\setlength{\parskip}{0pt}
Relying solely on training loss can blindside data allocation. As shown in Table ~\ref{tab:proposal}, while target–prediction KL decreased monotonically across both stages (2.19 $\rightarrow$ 0.66 $\rightarrow$ 0.64), net decision accuracy surged ($+$78 correct) before unexpectedly reversing ($-$13). Capturing this divergence allowed our orchestrator to dynamically throttle the bucket's mixture weight (5.36\% $\rightarrow$ 5.14\%).

%Table~\ref{tab:proposal} illustrates this connection on 341 fixed data-processing and agent-workflow questions with hard targets. KL fell in both comparisons, while net correctness first increased by 78 and then decreased by 13. Feedback from this group entered a joint update across task buckets, producing the next-stage mixture shares shown alongside.
\end{minipage}
\end{lrbox}%
\usebox{\workflowfeedbacktextbox}\hfill%
\begin{minipage}[t][\dimexpr\ht\workflowfeedbacktextbox+\dp\workflowfeedbacktextbox-2.5pt\relax][s]{0.53\linewidth}
\vspace{0pt}
\small
\setlength{\parskip}{0pt}
\captionsetup{justification=raggedright,singlelinecheck=false,skip=4pt,aboveskip=0pt}
\caption{Workflow feedback and mixture updates, across 341 fixed data-processing and agent-workflow questions with hard targets.}
\label{tab:proposal}
\vfill
\setlength{\tabcolsep}{3pt}
\renewcommand{\arraystretch}{1.16}
\begin{tabular*}{\linewidth}{@{\extracolsep{\fill}}lrrr@{}}
\toprule
\textbf{\color{picink}Stage} & \textbf{\color{picink}Mean KL} & \textbf{\color{picink}$\Delta$ correct} & \textbf{\color{picink}Mixture share (\%)} \\
\midrule
Earlier & $2.19\rightarrow0.66$ & $+78$ & $4.69\rightarrow5.36$ \\[2pt]
Later & $0.66\rightarrow0.64$ & $-13$ & $5.36\rightarrow5.14$ \\
\bottomrule
\end{tabular*}
\par\vfill
{\footnotesize $\Delta$ correct: net change on the same 341 questions. Share: change in mixture weight after the joint update across task buckets.\par}
\end{minipage}
\end{table}

\paragraph{Persistent Assets and Research Memory.}
To preserve experimental context, we link proposals, data, models, configurations, protocols, and results. We apply DataJuicer-Panel to manage these records within $\mathcal R_t$. Standardized records support cross-branch retrieval; lineage tracks model ancestry and reused recipes or findings. Memory associates gains and regressions with slices and settings, guiding experiments from other starting models.

\paragraph{Composable Experiment Execution.}
To execute a proposal, we compose data, training, inference, and evaluation steps. We use DataJuicer-Sandbox~\cite{sandbox}: \emph{Factories} provide these capabilities, \emph{Hooks} bind assets and settings, and \emph{Pipelines} sequence steps. These interfaces combine the data operations in Figure~\ref{fig:rsi} under the settings in Table~\ref{tab:boundaries}; analysis and execution proceed concurrently.

\begin{table}[H]
\centering\small
\caption{Data, training, evaluation, and research settings: fixed within comparisons, adaptable across proposals.}
\label{tab:boundaries}
\setlength{\tabcolsep}{6pt}
\renewcommand{\arraystretch}{1.12}
\begin{tabularx}{\linewidth}{@{}p{.15\linewidth}YY@{}}
\toprule
\textbf{\color{picink}Component} & \textbf{\color{picink}Fixed within a comparison} & \textbf{\color{picink}Varied between comparisons} \\
\midrule
Data & Pool version, group IDs, admission rules & Selection, mixtures, synthesis, new sources \\
Training & Parent weights, targets, optimizer, schedule & Parent, objective, learning rate, merging \\
Evaluation & Questions, scoring, inference settings & Input arrangement, diagnostic cohorts \\
Research & Target, comparator, devices and time & Analysis tools, priorities, follow-ups \\
\bottomrule
\end{tabularx}
\end{table}

\paragraph{Validation and Comparison.}
\label{sec:rsi-heldout}
Development comparisons track improvements and retained skills; disjoint confirmation sets and broad evaluations assess transfer. Source and construction identities organize splits, while semantic labels expose coverage. Audits check sample validity, input versions, and evaluation completeness. Outcomes and execution costs return to the shared research state (Appendix~\ref{app:feedback_validation}).

\paragraph{Proposal Selection and Human Steering.}
Agents rank proposals by expected improvement, retained skills, information value, and cost, recording their target, parent, action, and comparator. Short diagnostics resolve uncertainty; longer runs develop promising directions. Human researchers set priorities, review major changes, and introduce new methods and data sources to be explored. The human-preference and agent-accumulated findings collectively determine the next experiments.
\par\endgroup

\subsection{Instantiation in Piccolo Development}
\label{sec:rsi-instantiation}

\begin{figure}[!ht]
\centering
\includegraphics[width=\linewidth]{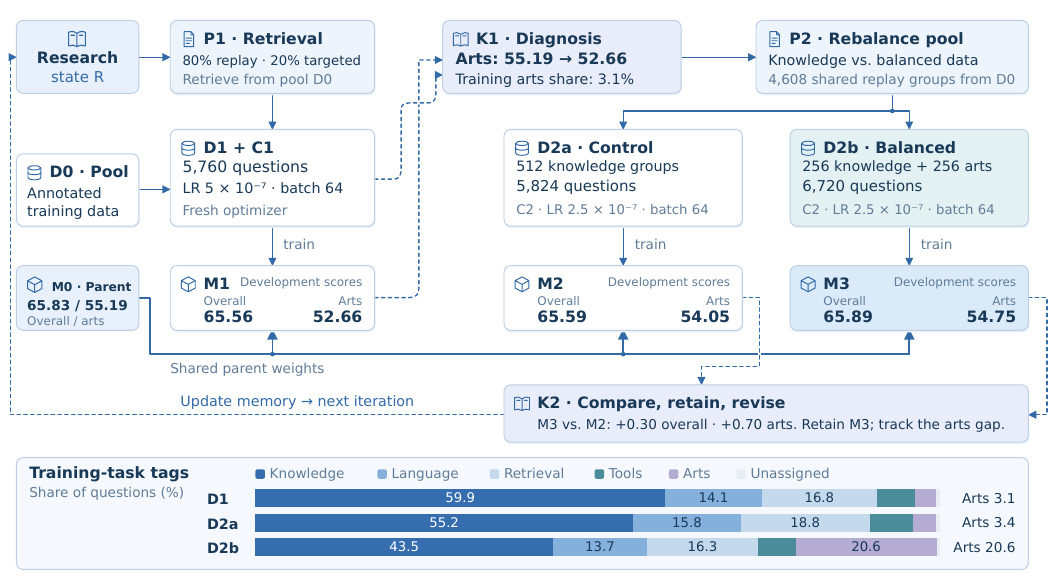}
\caption{A data-rebalancing case from Piccolo training. \textbf{D}: Data; \textbf{C}: Training Configuration; \textbf{P}: Proposal; \textbf{M}: Model; \textbf{K}: Iteration Knowledge. Shared-parent branches compare knowledge-only and balanced additions after an arts regression. Bars show task distributions; balanced additions yield higher overall and arts scores.}
\label{fig:iteration-case}
\end{figure}

We instantiated the orchestrator with a retained model, an indexed data pool, a development protocol, and a compute allocation. Training runs returned models, predictions, measured cost, and findings for subsequent proposals. Figure~\ref{fig:iteration-case} traces an effective data-rebalancing iteration from actual Piccolo training, showing how the assets, execution, and memory components in Section~\ref{sec:rsi-design} support successive proposals.

\textbf{(1) Execute a retrieval proposal.} P1 combined replay with a retrieved supplement to train M1 from retained model M0. \textbf{(2) Diagnose the regression.} Arts performance fell from 55.19 to 52.66; joining these predictions to the selected-data records showed an arts share of 3.1\% among training questions. Diagnosis K1 made arts coverage the focus of the next experiment. The asset records preserved the selected groups, configuration, and parent model alongside this finding.

\textbf{(3) Revise the mixture and compare.} P2 tested knowledge-only and balanced knowledge--arts supplements from the same M0, with shared replay groups, learning rate, and seed. Complete-group selection preserved native supervision: recipe shares count groups, tag shares count questions, and group composition determines question and token totals. Under the common development protocol, balanced model M3 outscored M2 by 0.30 points overall and 0.70 on arts. Memory K2 retained this comparison and the remaining arts gap to M0 for subsequent proposals. Model lineage records the shared parent; research lineage carries the diagnosis from P1 into P2 and its findings back into $\mathcal R_t$.

%% file: sections/04_evolution.tex
\clearpage
\section{Performance and Model--Data Co-Evolution}
\label{sec:exploration}

%Piccolo development couples two evolving objects: the data recipe and the decision model that learns from it. We first report family-level performance, then follow recorded branches of data and model evolution, connecting their proposals, updates, and resulting capabilities.
Post-training is traditionally treated as optimizing a dynamic model over a static data recipe. 
Piccolo reframes data-model development as a coupled dynamical system: as the model’s capability boundary shifts, the optimal data distribution and recipe \textbf{must} shift with it. A recipe that accelerates learning in an early checkpoint or on one backbone generation can induce regressions in another. 
% We first report family-level performance along the size–quality frontier, and then trace how co-evolving data allocations alongside model architectures resolves capability trade-offs that neither can solve alone.

\subsection{Overall Performance}
\label{sec:overall-results}

\begin{table}[!htb]
\centering\small
\caption{Vision and application scores for the same model families: (a) Decision Index Vision; (b) HarnessJevBench, DataJevBench, and JevBench. Bold/underlined scores mark the best/second-best per column. Full averages Public and Private; Harness uses hard targets. ``---'' denotes a pending submission.}
\label{tab:applications}
\begin{minipage}[t]{.49\linewidth}
\vspace{0pt}\centering\footnotesize
\textbf{(a) Decision Index Vision}\par\vspace{4pt}
\setlength{\tabcolsep}{1.5pt}
\renewcommand{\arraystretch}{1.08}
\begin{tabular*}{\linewidth}{@{\extracolsep{\fill}}lrrr@{}}
\toprule
\textbf{Model} & \textbf{Full} & \textbf{Private} & \textbf{Public} \\
\midrule
\textcolor[HTML]{3C5578}{\textbf{Piccolo-27B}} & \textbf{75.59} & \textbf{70.97} & \underline{80.22} \\
JDI-Decision V1 & \underline{75.05} & \underline{70.52} & 79.58 \\
WeMM-JEV-27B & 73.81 & 69.70 & 77.93 \\
Valen 1.5 27B & 73.29 & 63.68 & \textbf{82.90} \\
Sieve 31B v3 & 72.07 & 68.57 & 75.57 \\
StartLux v1.1 27B & 71.61 & 69.39 & 73.83 \\
\midrule
\textcolor[HTML]{3C5578}{\textbf{Piccolo-9B}} & 68.35 & 65.04 & 71.67 \\
Valen 1.5 9B & 69.67 & 59.34 & 80.00 \\
D3 Flash 9B & 66.89 & 63.91 & 69.87 \\
\textcolor[HTML]{3C5578}{\textbf{Piccolo-4B}} & 67.46 & 63.49 & 71.42 \\
Valen 1.5 4B & 66.58 & 55.35 & 77.81 \\
D3 Mini 4B & 63.16 & 58.65 & 67.67 \\
\midrule
\textcolor[HTML]{3C5578}{\textbf{Piccolo-2B}} & --- & --- & 66.64 \\
\textcolor[HTML]{3C5578}{\textbf{Piccolo-0.8B}} & --- & --- & 55.17 \\
\bottomrule
\end{tabular*}
\end{minipage}\hfill
\begin{minipage}[t]{.49\linewidth}
\vspace{0pt}\centering\footnotesize
\textbf{(b) Application evaluations}\par\vspace{4pt}
\setlength{\tabcolsep}{1.5pt}
\renewcommand{\arraystretch}{1.08}
\begin{tabular*}{\linewidth}{@{\extracolsep{\fill}}lrrr@{}}
\toprule
\textbf{Model} & \textbf{Harness} & \textbf{Data} & \textbf{JevBench} \\
\midrule
\textcolor[HTML]{3C5578}{\textbf{Piccolo-27B}} & \textbf{84.41} & \underline{82.60} & \underline{92.21} \\
JDI-Decision V1 & 82.71 & \textbf{84.40} & 90.48 \\
WeMM-JEV-27B & 82.03 & 82.40 & \underline{92.21} \\
Valen 1.5 27B & 79.15 & 80.20 & 91.34 \\
Sieve 31B v3 & \underline{83.05} & 80.00 & 90.48 \\
StartLux v1.1 27B & 82.03 & 79.60 & \textbf{93.94} \\
\midrule
\textcolor[HTML]{3C5578}{\textbf{Piccolo-9B}} & 71.53 & 72.40 & 88.74 \\
Valen 1.5 9B & 64.58 & 66.40 & 86.15 \\
D3 Flash 9B & 65.42 & 68.20 & 82.25 \\
\textcolor[HTML]{3C5578}{\textbf{Piccolo-4B}} & 66.10 & 70.20 & 87.01 \\
Valen 1.5 4B & 62.03 & 65.40 & 83.55 \\
D3 Mini 4B & 65.93 & 65.80 & 81.82 \\
\midrule
\textcolor[HTML]{3C5578}{\textbf{Piccolo-2B}} & 56.10 & 52.80 & 86.15 \\
\textcolor[HTML]{3C5578}{\textbf{Piccolo-0.8B}} & 51.19 & 52.20 & 75.32 \\
\bottomrule
\end{tabular*}
\end{minipage}
\end{table}

\paragraph{Decision Index.}
Piccolo-27B ranks first on Vision \textbf{Full (75.59) and Private (70.97)}, with 80.22 Public (Table~\ref{tab:applications}a). Its Public profile exceeds JDI-Decision V1 by \textbf{1.78 points on CV-Bench, 4.02 on BLINK, and 3.38 on MMMU-Pro Vision}; Appendix~\ref{app:vision-profiles} gives all eleven tasks. Piccolo-9B and 4B score 65.04 and 63.49 on Private, while 2B and 0.8B score 66.64 and 55.17 on Public.

\paragraph{Application evaluations.}
HarnessJevBench covers tool use, permissions, memory, and execution; DataJevBench covers filtering, processing plans, and output quality; JevBench covers general decisions. Piccolo-27B scores 84.41 on Harness, 82.60 on Data, and 92.21 on JevBench, outperforming the strongest external Harness baseline by 1.36 points in the completed batch. Piccolo-9B and 4B score \textbf{71.53/66.10 on Harness} and \textbf{72.40/70.20 on Data}, exceeding both size-matched D3 and Valen models. Their Private Vision scores also exceed both baselines. Appendix~\ref{app:application-series} gives the full comparison.

\begin{figure}[!htb]
\begin{minipage}[t]{.43\linewidth}
\vspace{4pt}
\textbf{Capability profiles.}
Piccolo-4B, 9B, and 27B all lie on the \textbf{Private size--quality frontier} (Figure~\ref{fig:pareto}). The 4B model trades 1.55 Private points for fewer than half the served parameters of 9B (63.49 vs. 65.04). With roughly one-third the parameters of 27B, the 9B model scores 5.93 points lower on Private (65.04 vs. 70.97). The largest model extends the frontier to 70.97. These operating points offer different capacity budgets for the same candidate-scoring interface; the task profiles identify which capabilities each deployment retains.
\end{minipage}\hfill
\begin{minipage}[t]{.50\linewidth}
\vspace{0pt}\centering
\includegraphics[width=0.9\linewidth]{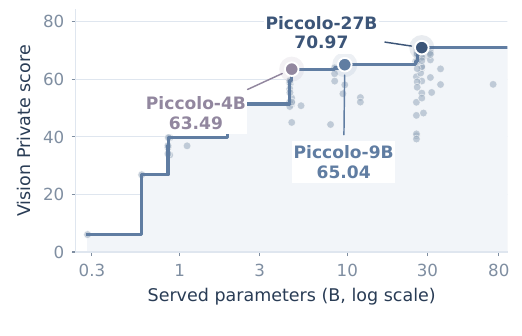}
\caption{Vision Private size--quality frontier. Points use reported served parameters; the frontier follows the best score available within each parameter budget.}
\label{fig:pareto}
\end{minipage}
\end{figure}

\clearpage
\subsection{Data Evolution}
\label{sec:data-proposals}

Data evolution involves \emph{coverage} (D1), \emph{allocation} (D2), and \emph{joint recipes} (D3).

\begin{figure}[!ht]
\centering
\includegraphics[width=\linewidth]{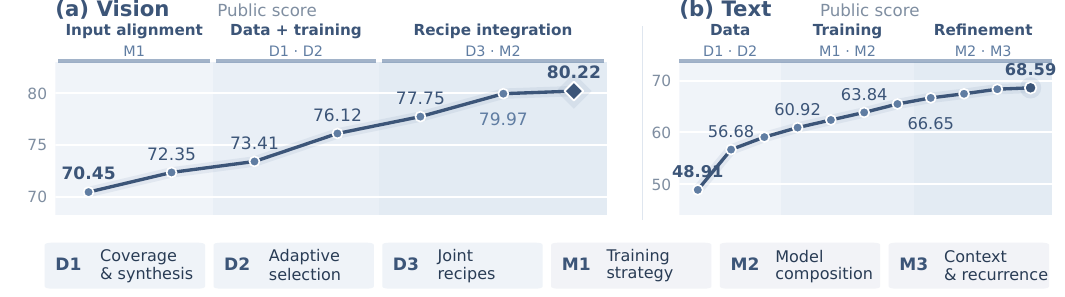}
\caption{Public exploration milestones and research directions. Stage bands group the exploration periods; data directions (D1--D3) and model directions (M1--M3) connect the milestones to coverage, training, composition, and inference developments. D1--D3 index data evolution in Section~\ref{sec:data-proposals}; M1--M3 index model evolution in Section~\ref{sec:model-evolution}.}
\label{fig:evolution-overview}
\end{figure}

\paragraph{\hyperref[fig:evolution-overview]{D1}\quad Feedback-guided data revision restores coverage.}
Held-out feedback and semantic labels identify gaps for \textbf{\textit{targeted replacement}}: replacing 10\% of an early recipe outperforms matched random replacement by 2.11 development points (56.56 vs. 54.45). Later, restoring underrepresented \textbf{\textit{preference coverage}} raised the arts/preference component from 48.91 to 53.42. For sparse slices, proposals specify automated collection with deduplication and sensitive-information removal, or \textbf{\textit{model-guided synthesis}} from verified seeds. The model's pass rates on the new tasks---63.43\% for state tracking and 15.38\% for budget allocation---then identified where difficult examples remained.

\begin{figure}[!htb]
\centering
\includegraphics[width=\linewidth]{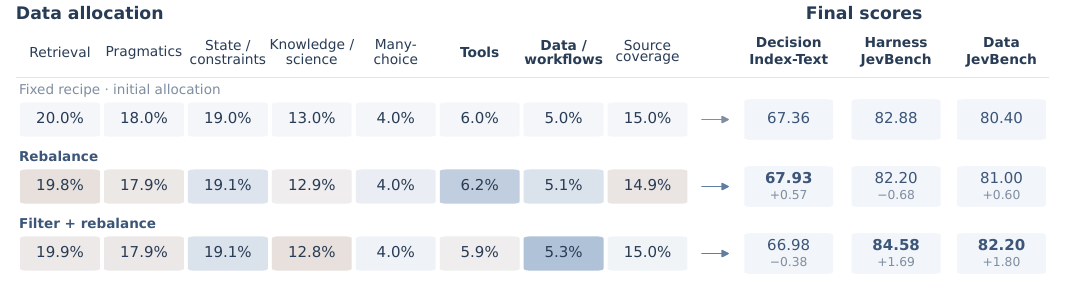}
\caption{Data allocations and outcomes across a fixed recipe, feedback-guided rebalancing, and rebalancing with additional filtering. Blue/tan cells mark allocation increases/decreases from the fixed recipe. Right-hand cells report final scores and signed changes from that reference. \takeaway{Rebalancing gives the strongest Text result; adding filtering improves Harness and Data, producing distinct capability profiles from different data actions.}}
\label{fig:adaptive-data}
\end{figure}

\paragraph{\hyperref[fig:evolution-overview]{D2}\quad Training dynamics guide adaptive data allocation.}
The \emph{Structured Feedback} interface in Section~\ref{sec:rsi-design} supplies target--prediction KL and changes in decision correctness on held-out questions. These readouts guide \emph{Filter \& Rebalance} in Figure~\ref{fig:rsi}: \textbf{\textit{rebalancing}} revises bucket shares as learning progresses, while \textbf{\textit{filtering}} selects examples using source-level loss and verification outcomes. Figure~\ref{fig:adaptive-data} aligns the resulting allocations with each branch's scores. Rebalancing elevated Public Text to 67.93, versus 67.36 for the fixed recipe. Adding filtering reached \textbf{84.58 Harness and 82.20 Data}, with Text at 66.98.

\begin{figure}[!tb]
\centering
\includegraphics[width=\linewidth]{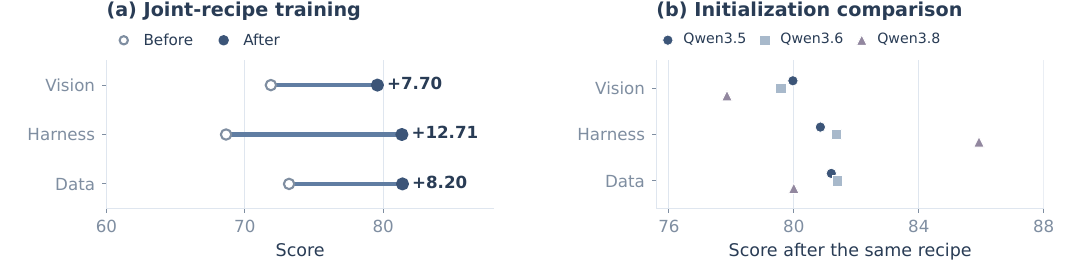}
\caption{Recipe--model interaction. (a) Qwen3.6-27B before and after joint training. (b) Three 27B initializations trained with the same recipe and data order. Vision scores use the Public suite.
\takeaway{Joint recipes deliver broad gains across tasks (a), while the optimal backbone depends heavily on the target deployment domain (b).}}
\label{fig:recipe_gains}
\end{figure}

\paragraph{\hyperref[fig:evolution-overview]{D3}\quad Joint recipes consolidate gains from data evolution.}
\textbf{\textit{Joint decision training}} combined general decisions, image-dependent tasks, and workflow additions, with replay material for established skills. \textbf{\textit{OCR-aware SFT}} aligned supervision with image-associated text available at inference. At one consolidation node, a Qwen3.6-27B descendant improved from \textbf{71.89 to 79.59} on Vision, \textbf{68.64 to 81.36} on Harness, and \textbf{73.20 to 81.40} on Data (Figure~\ref{fig:recipe_gains}a). 
%The retained recipe also enabled \textbf{\textit{initialization selection}} under a common data order (Figure~\ref{fig:recipe_gains}b): Qwen3.5 led Vision at 79.97, while Qwen3.8 reached \textbf{85.93 Harness}, versus 80.85 and 81.36 for Qwen3.5 and Qwen3.6. The same data recipe thus supported different deployment priorities through backbone choice.
Crucially, newer backbones do not guarantee across-the-board superiority under a fixed recipe (Figure~\ref{fig:recipe_gains}b). 
Trained on the exact same data order, the earlier Qwen3.5 initialization (79.97) outperformed both Qwen3.6 and Qwen3.8 on Vision (79.59 and 77.86), whereas Qwen3.8 dominated agentic Harness tasks (85.93), versus 80.85 and 81.36 for Qwen3.5 and Qwen3.6. Model initialization and data recipes must therefore be co-adapted to match target deployment priorities.

\FloatBarrier

\subsection{Model Evolution}
\label{sec:model-evolution}

Model proposals reused these data assets to compare training strategies (M1), model composition and refinement (M2), and inference (M3), linking each choice to its measured capability response.

\paragraph{\hyperref[fig:evolution-overview]{M1}\quad Training choices benefit from matched supervision.}
\textbf{\textit{Bidirectional attention}} benefited from matched training and inference: Text development scores reached \textbf{69.01}, versus 68.20 with causal attention and 68.14 when only inference became bidirectional. \textbf{\textit{Staged SFT and RL}} paired broad supervision with targeted adaptation. Full-parameter and low-rank continuations tested how broadly to update the model. OCR-aware SFT established the input--decision correspondence before candidate-level RL and supervised replay addressed residual errors. \textbf{\textit{Objective and data weighting}} then evolved together: from a shared parent and the same examples, fixed-weight GRPO reached 77.13 Public Vision, candidate CE 77.31, and confidence/progress-weighted GRPO \textbf{77.35}. These results made both the supervision format and the distribution of learning weight part of the training proposal.

\Needspace{4\baselineskip}
\paragraph{\hyperref[fig:evolution-overview]{M2}\quad Model composition integrates complementary capabilities.}
\label{par:model-composition}
\textbf{\textit{Cross-modal merging}} combined branches strengthened by different data recipes. In one exploration, retaining the visual modules of an image--text model and blending its language parameters with a text-oriented branch boosted Public Text from 67.89 to 68.15 while trading off Vision from 77.29 to 75.70 (Table~\ref{tab:model-composition}a). The mixture retained part of each branch's advantage. \textbf{\textit{Candidate-distribution integration}} explored a complementary route inspired by MOPD~\cite{mopd}. Reference predictions supplement the original targets on selected examples. Exact matching reached \textbf{82.80 Data}, versus 82.20 for the original-target continuation, while Harness stayed at 84.92 and Vision remained near 77.07 (Table~\ref{tab:model-composition}b). These branches provide distinct task profiles for subsequent composition.

\begin{table}[!htb]
\centering\small
\caption{Capability integration through (a) language-weight interpolation with visual modules retained and (b) training with reference candidate distributions. Rows compare each parent with its merged or further-trained descendants across tasks. Text and Vision use Public scores.}
\label{tab:model-composition}
\begin{minipage}[t]{.42\linewidth}
\vspace{0pt}\centering\small
\textbf{\color{picink}(a) Cross-modal merging}\par\vspace{5pt}
\setlength{\tabcolsep}{3pt}
\renewcommand{\arraystretch}{1.17}
\begin{tabular*}{\linewidth}{@{\extracolsep{\fill}}lrr@{}}
\toprule
\textbf{Model / text share} & \textbf{Text} & \textbf{Vision} \\
\midrule
Text-oriented parent & \textbf{68.53} & 73.16 \\
Image--text parent & 67.89 & \textbf{77.29} \\
Merge: 25\% text & 67.94 & \underline{76.92} \\
Merge: 50\% text & \underline{68.15} & 75.70 \\
\bottomrule
\end{tabular*}
\end{minipage}\hfill
\begin{minipage}[t]{.56\linewidth}
\vspace{0pt}\centering\small
\textbf{\color{picink}(b) Candidate-distribution integration}\par\vspace{5pt}
\setlength{\tabcolsep}{2pt}
\renewcommand{\arraystretch}{1.17}
\begin{tabular*}{\linewidth}{@{\extracolsep{\fill}}lrrr@{}}
\toprule
\textbf{Training targets} & \textbf{Vision} & \textbf{Harness} & \textbf{Data} \\
\midrule
Parent & \textbf{77.38} & \underline{84.24} & \underline{82.60} \\
Original targets & \underline{77.09} & \textbf{84.92} & 82.20 \\
+ exact matching & 77.07 & \textbf{84.92} & \textbf{82.80} \\
+ sampled matching & 77.07 & \textbf{84.92} & 82.40 \\
\bottomrule
\end{tabular*}
\end{minipage}
\end{table}

\textbf{\textit{Weight-space editing}} refined candidate combinations through module scaling and low-rank update directions~\cite{evolutionarymerge}. One selected configuration combined \emph{(i)} approximately 5\% stronger visual attention projections in blocks 9--17, \emph{(ii)} slight attenuation of feed-forward outputs in language layers 8--23, \emph{(iii)} log-decay offsets in selected late gated-attention blocks, and \emph{(iv)} rank-8 updates to output projections in layers 30--31. The combined model elevated Public Vision to 70.75 (vs. 70.10 for the retained reference), with a 0.68-point Harness decrease. Earlier, merging five data-specialized descendants boosted Public Text to 63.11, surpassing the strongest individual branch at 62.78.

\begin{figure}[!htb]
\centering
\includegraphics[width=\linewidth]{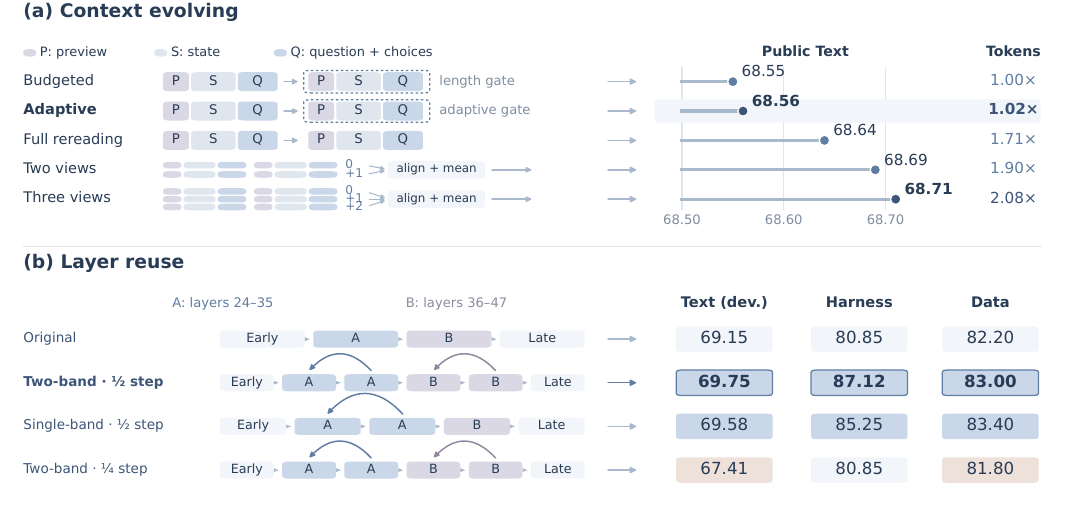}
\caption{Inference choices on fixed 27B models. (a) Context layouts: Public Text scores and input tokens relative to budgeted rereading. (b) Layer reuse: Text development, Harness, and Data scores. Colored blocks depict input arrangements or repeated layer bands. \takeaway{Input expansion yields small gains at higher token cost; layer reuse is sensitive to step size.}}
\label{fig:recurrent-inference}
\end{figure}

\Needspace{9\baselineskip}
\paragraph{\hyperref[fig:evolution-overview]{M3}\quad Input organization and layer reuse offer distinct inference gains.}
\textbf{\textit{Context evolving}} changed question previews, repeated context, and aggregation across candidate-label views (Figure~\ref{fig:recurrent-inference}a). Adaptive rereading reached \textbf{68.56 Public Text at 1.02$\times$ input volume}. Full rereading scored 68.64; two- and three-view aggregation reached 68.69 and 68.71 at 1.90$\times$ and 2.08$\times$ input volume.

\textbf{\textit{Recurrent inference}} reused layer ranges within the forward pass (Figure~\ref{fig:recurrent-inference}b). Two half-step passes over each of two adjacent ranges raised Text development scores from \textbf{69.15 to 69.75} and Harness from \textbf{80.85 to 87.12}. Reusing a single range yielded 69.58 on Text, whereas reducing the two-range step to one quarter lowered the score to 67.41. A separate recurrent configuration raised 2B Public Vision from 65.85 to 66.64.

%% file: sections/05_discussion.tex
\FloatBarrier
\section{Discussion}
\label{sec:discussion}

%\subsection{Validation Coverage and Transfer}
\subsection{Optimizing Single Benchmarks Masks Cross-Task Regressions}
\label{sec:transfer-discussion}

\begin{figure}[!htbp]
\centering
\includegraphics[width=\linewidth]{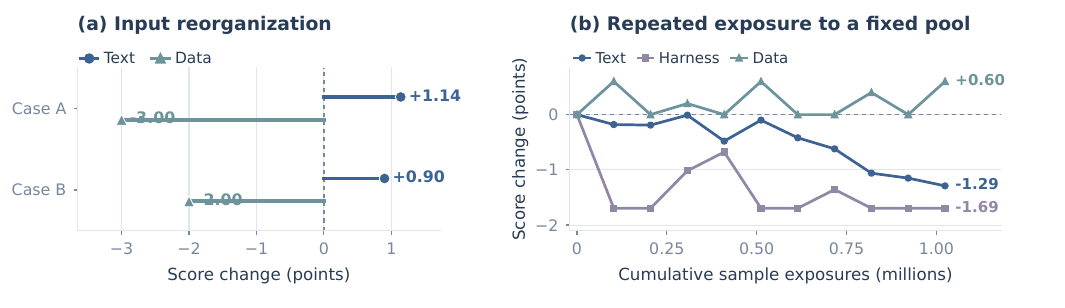}
\caption{Cross-task responses to (a) input reorganization at fixed weights and (b) repeated training on a fixed pool. Zero marks the starting score; positive and negative values indicate gains and regressions. \takeaway{Gains and regressions coexist across tasks, making cross-task validation central to proposal selection and the choice between further training and revised data.}}
\label{fig:transfer-exposure}
\end{figure}

%\textbf{A gain in one task family can accompany a regression in another.} With model weights fixed, one input-layout change raised Public Text from 65.05 to 66.19 while Data fell from 84.4 to 81.4. A second model showed the same direction (Figure~\ref{fig:transfer-exposure}a). 
\textbf{Isolated gains in one capability often mask severe regressions in another, even without weight updates.} Merely reorganizing the input layout at fixed weights boosted Public Text by +1.14 points while causing a 3.00-point drop on DataJevBench (Figure~\ref{fig:transfer-exposure}a).
Semantic slices locate the affected capabilities; held-out task and source groups assess their transfer (Section~\ref{sec:rsi-heldout}). This profile identifies the skills to preserve or strengthen in subsequent data revisions.

%\subsection{Data and Compute Efficiency}
\subsection{More Compute on Fixed Data Degrades Generalization}
\label{sec:efficiency-discussion}

%\textbf{Further exposure changes the balance of learned capabilities.} Across eleven recorded checkpoints on a fixed pool, 1.024 million sample exposures changed Text from 65.13 to 63.84 and Harness from 83.22 to 81.53, while Data rose from 84.0 to 84.6 (Figure~\ref{fig:transfer-exposure}b). 
\textbf{Scaling training exposures on a static data pool skews the capability balance rather than uniformly improving the model.} Over 1M sample exposures on a fixed pool, Data scores crept up by $+$0.60, but Text and Harness degraded significantly by $-$1.29 and $-$1.69 points (Figure~\ref{fig:transfer-exposure}b). 
This demonstrates that dynamic data curation is more compute-efficient than prolonged static training.
In addition, the diverging trajectories motivate comparing longer training with revised coverage and allocation. Proposals from a shared retained model make these alternatives directly comparable.

Inference introduces another allocation choice. Adaptive rereading achieved 68.56 Public Text; three-view aggregation further reached 68.71 with \textbf{2.04$\times$ the input tokens} of adaptive rereading. Structural changes were similarly sensitive to their setting: two-band half-step recurrence gained 0.60 Text development points, whereas quarter-step recurrence lost 1.74 (Figure~\ref{fig:recurrent-inference}). Head- and channel-level rescaling also entered the search; strong visual-branch attenuation failed to improve the retained development score. The useful computation depends on both the task and the current model.

\subsection{Capability Integration and Research Reuse}
\label{sec:research-reuse}

\textbf{Combining specialized branches changes which capabilities survive.}
The cross-modal merge in Table~\ref{tab:model-composition}a gained 0.26 Public Text points over its image--text parent while losing 1.59 Vision points. Candidate-distribution matching produced a smaller redistribution: Data improved by 0.60 over training on the original targets, with Harness unchanged and Vision nearly unchanged (Table~\ref{tab:model-composition}b). Branch selection can use complementary task strengths, with the retained gains measured after integration.

%An unsuccessful branch can contribute a useful \textbf{parameter update}. In one recorded lineage, a continuation scored below its parent on the development set; a later successful combination reused selected components of that update with a negative coefficient. 
Even ``failed'' training runs contain actionable gradient signals. In one notable lineage, a continuation branch regressed below its parent on the development set; however, extracting specific components of its parameter update and applying them with a \textbf{negative coefficient} yielded a winning weight-space merge.
%Recording the parent, update, and task-level response makes these combinations recoverable for subsequent proposals (Appendix~\ref{app:discussion-records}).
%This shows that in an agentic research loop, preserving the full experimental lineage helps. By connecting parent checkpoints, parameter deltas, and slice-level responses, the orchestrator can convert apparent dead-ends into composable building blocks for future iterations (Appendix~\ref{app:discussion-records}).
This underscores the value of preserving full experimental lineage: by connecting parent checkpoints, parameter deltas, and slice-level responses, the orchestrator converts apparent dead-ends into composable building blocks for future iterations (Appendix~\ref{app:discussion-records}).

\subsection{Relation to Prior Work}
\label{sec:related-work}

Decision-model research explores complementary training and inference choices. Mammoth and Elephant combine data recipes and bidirectional attention~\cite{mammoth}; Drex studies supervised and reinforcement learning~\cite{drex}; StartLux adds confidence-guided deliberation~\cite{startlux}. Evolutionary merging and MOPD study capability integration~\cite{evolutionarymerge,mopd}. Agentic Data Engineering~\cite{ade} and open-ended self-improvement~\cite{dgm} guide research through feedback and accumulated discoveries. Piccolo connects these ideas through candidate-level feedback, evolving data recipes, and reusable model branches.

%% file: sections/06_outlook.tex
\FloatBarrier
\section{Conclusion and Future Directions}

Piccolo demonstrates that advancing frontier decision models requires more than scaling parameters or static datasets; it requires scaling the research loop itself. Through Agentic Data Engineering, Piccolo co-evolves data curricula and model weights across a 0.8B to 27B family, establishing state-of-the-art performance across visual judgment, agent harnesses, data workflows, and interactive control.

Beyond the released weights, our development campaign exposes fundamental dynamics of post-training: surrogate loss optimization frequently decouples from discrete decision accuracy, static data pools inevitably induce cross-capability interference, and optimal data recipes are strictly contingent on backbone initialization. By embedding these diagnostics into an autonomous, memory-backed orchestration loop, we shift post-training from manual trial-and-error to systematic, self-correcting engineering.

Looking forward, the convergence of fast System-1 decision models and deliberative System-2 reasoning points toward hybrid cognitive architectures. Future work will extend this recursive co-evolution to adaptive test-time deliberation, multi-turn environmental interaction, and automated synthesis for long-tail capability frontiers; ultimately advancing not just the models that make decisions, but the autonomous systems that build them.

%% file: sections/team.tex
\hypertarget{piccolo-team}{}
\pdfbookmark[1]{Piccolo Team}{piccolo-team-bookmark}
\section*{Piccolo Team}
\label{team:contributions}
\noindent \textbf{Model:} Zhenqing Ling, Qirui Jiao, Daoyuan Chen; \\ \textbf{Data:} Cheng Chen, Xinji Mai, Haowen Gao; \\ \textbf{Infra:} Zhuo Zhang, Xuchen Pan, Mengsu Ding, Jiaji Deng, Yuexiang Xie, Xiaoming Qin, Weirui Kuang; \\ \textbf{Project Steering:} Zhaoyang Liu, Yaliang Li, Bolin Ding, Jingren Zhou.\par

%% file: sections/appendix/full_public_rankings.tex
\FloatBarrier
\section{Full and Public Vision Rankings}
\label{app:full_public_rankings}
\hypertarget{full-public-rankings}{}

The Full-score view complements the Private comparison in Figure~\ref{fig:public_vision}. The two panels retain the fifteen highest Full scores overall and among nominal parameter classes below 27B. Full is the equal-weight mean of Public and Private; dashed outlines show the difference to Public. Both views use the leaderboard snapshot specified in the opening footnote \ref{fig:public_vision}. Table~\ref{tab:full_public} reports the three Piccolo score components.

\begin{figure}[!htbp]
\centering
\includegraphics[width=\linewidth]{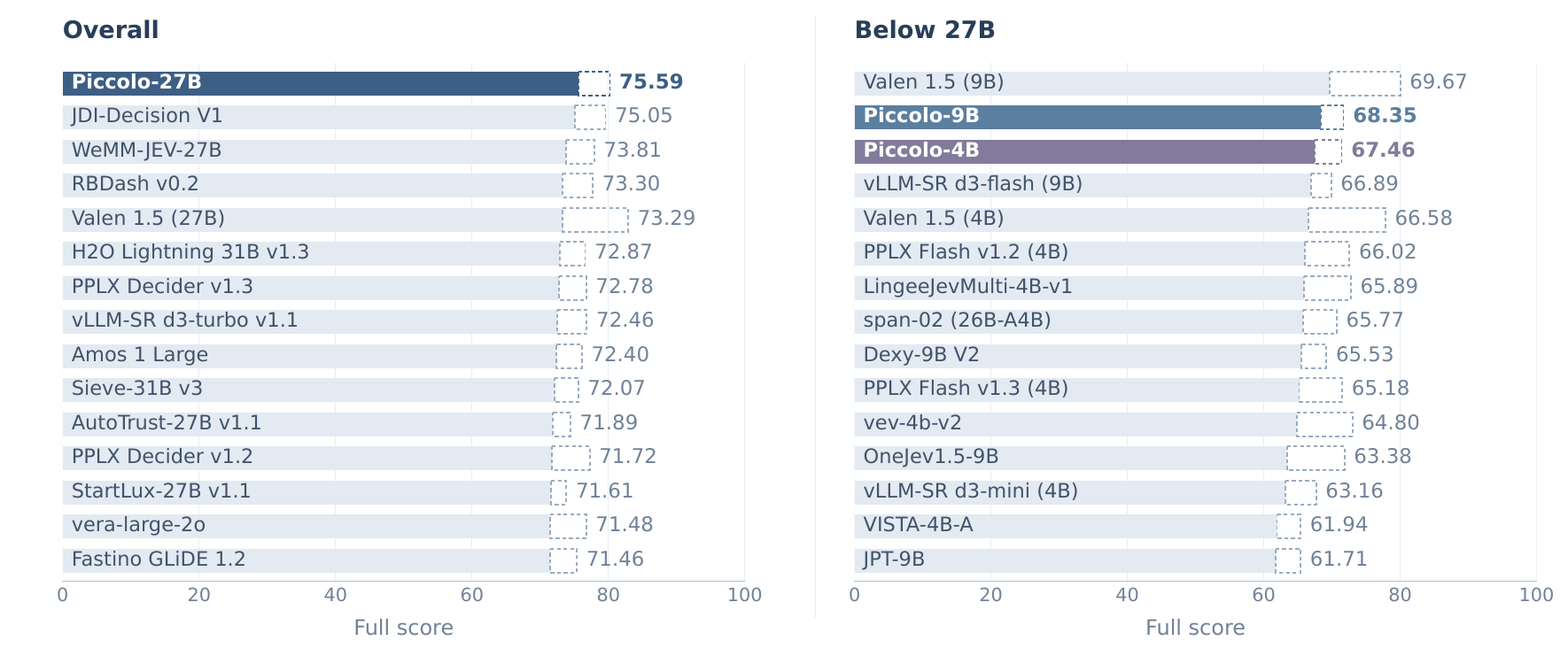}
\caption{Decision Index Vision Full rankings: top fifteen overall (left) and below 27B (right). Filled bars and numerical labels report Full scores; dashed outlines span the difference between Public and Full.}
\label{fig:full_public_rankings}
\end{figure}

%% file: sections/appendix/candidate_supervision.tex
\FloatBarrier
\section{Candidate Supervision and Input Organization}
\label{app:candidate_supervision}

\subsection{Supervision and Decision Error}
\label{app:decision_format}

Section~\ref{sec:definition} defines the shared decision format and candidate readout in Equation~\ref{eq:candidate_readout}. Equation~\ref{eq:candidate_supervision} gives the supervised objective and its decomposition into target entropy and distributional mismatch. Here we relate this loss to decision error and detail the input implementation.

For hard-target decisions, loss also supplies a simple bound on the error indicator. With natural logarithms and $\hat y=\arg\max_k p_{\theta,k}$, an incorrect prediction implies $p_{\theta,y}\leq 1/2$, hence

\begin{equation}
\mathbf{1}[\hat y\ne y]\leq \frac{-\log p_{\theta,y}}{\log 2}.
\end{equation}

Averaging gives the same bound on empirical error by mean candidate loss divided by $\log 2$. At the population level, expected cross-entropy equals conditional target entropy plus expected KL to the true conditional distribution; minimizing the latter recovers the Bayes decision rule \cite{multiclasscalibration,multiclasslogloss}. In finite experiments, mean loss tracks probability changes across examples while accuracy tracks which candidates cross the decision boundary. Reporting both retains the graded training response and the operational answer changes, as in Table~\ref{tab:proposal}.

Hard-target correctness and target probability accompany this distributional measure. Slice comparisons retain supervision form, candidate count, source, and semantic labels. A repeated fixed-question measurement retains its input and target, while a training-window summary follows the examples encountered in that window.

Candidate scoring uses the native language-model head at the readout position. The compiler maps option order to the tokenizer's valid candidate labels, checks the mapping, and stores the target distribution separately from the visible prompt. Packed examples retain isolated boundaries. Bidirectional prompt masks apply to full-attention blocks; linear-attention blocks retain their original causal computation in the hybrid backbones.

Context evolving studies question-list prefixes, focused-question presentation, and length-aware layouts. Multimodal inputs preserve image order and question-to-image alignment. OCR text is supplied as an image-associated observation alongside the original image. A fixed recognizer prepares the training cache; evaluation extracts text from the input image using the corresponding preparation policy.

%% file: sections/appendix/data_recipes.tex
\FloatBarrier
\section{Data Sources and Training Overview}
\label{app:data_recipes}

\subsection{Source Inventory}
\label{app:source_inventory}

The initial text inventory contains 6.65M records and 8.93M questions from 43 source collections. A record can contain several questions. Table~\ref{tab:source_inventory} lists all 43 collections, grouped here by their primary use. Source conversion retains the original supervision and associates each question with its parent record. The visual inventory shown in Figure~\ref{fig:data} includes SAT-v2, Geo170K, TallyQA, NLVR2, TQA, and ArxivQA, covering spatial relations, geometry, counting, visual reasoning, diagrams, and scientific figures.

\begin{table}[!htbp]
\centering\small
\caption{The 43 collections in the initial text source inventory. Collection names retain their source identities; the groupings summarize their roles in data preparation.}
\label{tab:source_inventory}
\begin{tabularx}{\linewidth}{@{}>{\raggedright\arraybackslash}p{0.23\linewidth}Y@{}}
\toprule
\textbf{Primary use} & \textbf{Source collections} \\
\midrule
Language understanding, retrieval, and dialogue (10) & Banking77~\cite{data_banking77}, CLINC150~\cite{data_clinc150}, ESCI~\cite{data_esci}, GoEmotions~\cite{data_go_emotions}, iSarcasmEval~\cite{data_isarcasmeval}, MASSIVE~\cite{data_massive}, MNLI~\cite{data_mnli}, SGD~\cite{data_sgd}, SNLI~\cite{data_snli}, WANLI~\cite{data_wanli} \\
Knowledge and reasoning (10) & AQuA-RAT~\cite{data_aqua_rat}, ARC~\cite{data_arc}, BoolQ~\cite{data_boolq}, CommonsenseQA~\cite{data_commonsenseqa}, Cosmos QA~\cite{data_cosmos_qa}, FinQA~\cite{data_finqa}, HellaSwag~\cite{data_hellaswag}, MuSiQue Train~\cite{data_musique_train}, QASC~\cite{data_qasc}, WinoGrande~\cite{data_winogrande} \\
Preferences and quality judgments (5) & Bekko Selected~\cite{data_bekko_selected}, HelpSteer2~\cite{data_helpsteer2}, PRM800K~\cite{data_prm800k}, Prometheus~\cite{data_prometheus}, RAGTruth~\cite{data_ragtruth} \\
Actions and workflows (4) & JevForge Mind2Web~\cite{data_jevforge_mind2web,data_mind2web_original}, Procedural Typed Decisions~\cite{data_procedural_typed_decisions,data_reasoning_core}, ToolJudge Local, When2Call~\cite{data_when2call} \\
Converted and synthesized decision collections (14) & Eikos Decisions~\cite{data_eikos_decisions}, Helmo Synthetic~\cite{data_helmo_synthetic}, JEV Decisions Clean50K~\cite{data_jev_decisions_clean50k}, JEV Distill v3~\cite{data_jev_distill_v3}, MinoJev Data~\cite{data_minojev_data}, Open JEV~\cite{data_open_jev}, Open JEV v1.1~\cite{data_open_jev_v11}, Plumb Decisions~\cite{data_plumb_decisions}, Synthetic Typed Decisions~\cite{data_synthetic_typed_decisions}, TaskSource JEV~\cite{data_tasksource_jev,data_tasksource_original}, Typed Decisions~\cite{data_typed_decisions}, Typed Decisions JA~\cite{data_typed_decisions_ja,data_typed_decisions}, Typed Decisions PT/ES~\cite{data_typed_decisions_pt_es,data_typed_decisions}, Typed Decisions Synth~\cite{data_typed_decisions_synth} \\
\bottomrule
\end{tabularx}
\end{table}

\subsection{Data Composition}

Training curricula combine general text decisions, visual tasks, targeted synthesis, and replay of established skills. Figure~\ref{fig:data} shows a joint-pool snapshot of 663K examples: 311K text and 352K image--text decisions. Representative curricula draw on approximately 380K or 700K examples, with different balances of visual material and text coverage. Source and task-form metadata support sampling while preserving the original supervision distributions. Held-out feedback guides subsequent additions and allocation changes, as described in Section~\ref{sec:data-proposals}.

\subsection{Annotation Taxonomy and Semantic Buckets}
\label{app:semantic_taxonomy}

The annotation schema separates the subject of a decision from its required capability, task function, input quality, and estimated learning demands. Table~\ref{tab:taxonomy} lists its 13 axes and 551 selectable values. Domain has 334 leaf categories under 26 parent groups; capability has 144 leaf labels under 14 parent groups. The 40 parent groups organize the hierarchy, while the three native response interfaces---Choice, Noul, and Score---are stored separately as task-format metadata. Multiple labels can apply on axes such as decision basis or challenge factors.

\begin{table}[!htbp]
\centering\small
\caption{Semantic annotation vocabulary. Counts describe selectable values on each axis.}
\label{tab:taxonomy}
\begin{tabularx}{\linewidth}{@{}>{\raggedright\arraybackslash}p{0.24\linewidth}rY@{}}
\toprule
\textbf{Axis} & \textbf{Values} & \textbf{What it describes} \\
\midrule
Domain & 334 & Subject area, with 26 parent groups. \\
Capability & 144 & Required skill, with 14 parent groups. \\
Task function & 20 & The requested judgment, such as routing or tool-call validity. \\
Input quality & 3 & Usable input, missing information, or conflicting rules. \\
Verification method & 4 & Programmatic, semantic-model, hybrid, or manual checking. \\
Decision relation & 4 & Direct decision, evaluative judgment, decision support, or standalone solution. \\
Challenge factors & 11 & Structural demands such as interacting constraints or order sensitivity. \\
Input issues & 7 & Specific defects such as ambiguous targets or equivalent candidates. \\
Decision basis & 6 & Cost, time, benefit, risk, information gain, or rule compliance. \\
Objective structure & 5 & Feasibility, single objective, trade-off, inapplicable, or undefined criterion. \\
Knowledge requirement & 3 & Self-contained input, external knowledge, or insufficient input. \\
Difficulty & 5 & Levels 1–5 for the demands of answering the current question. \\
Learning value & 5 & Levels 1–5 for the learning signal suggested by the example's content. \\
\midrule
\textbf{Total} & \textbf{551} & \textbf{13 annotation axes.} \\
\bottomrule
\end{tabularx}
\end{table}

Annotations describe the input and requested decision. Supervision targets and measured outcomes remain separate fields joined by example identity. This supports comparisons within a domain, across capabilities, or under a fixed supervision form.

The semantic view in Figure~\ref{fig:data} covers approximately 210K annotated examples. Its 26 domain groups are summarized into everyday life (54.7\%), computing and technology (13.6\%), mathematics and science (12.7\%), humanities, arts, and society (5.8\%), and other domains (13.2\%). Choice, Noul, and Score account for roughly 59\%, 25\%, and 16\%; difficulty levels 2--3 account for about 77\%. These views describe the coverage available to feedback-guided selection.

\subsection{Training and Refinement Objectives}

Supervised stages optimize candidate cross-entropy using BF16 training, Adam, and a learning-rate schedule with warmup and decay. Proposals vary the data allocation and the trainable language, connector, and visual components. The comparisons in Figure~\ref{fig:recipe_gains} evaluate how a shared recipe interacts with backbone initialization.

Candidate-level reinforcement learning samples actions from the current candidate distribution and assigns correctness rewards. Group-normalized advantages $A_g$ guide the update, together with supervised anchoring and a frozen-reference distribution:
\begin{equation}
\ell_{\mathrm{target}}=-\frac{1}{G}\sum_{g=1}^{G}\operatorname{stopgrad}(A_g)\log p_\theta(a_g)
+\lambda\ell_{\mathrm{CE}}+\beta D_{\mathrm{KL}}(p_\theta\Vert p_{\mathrm{ref}}).
\end{equation}
Here $G$ is the number of sampled candidate actions, and $\lambda$ and $\beta$ control the supervised and reference terms. Supervised replay accompanies targeted updates. Weight-space proposals explore checkpoint interpolation, module rescaling, and low-rank update directions; held-out comparisons assess the resulting task profiles.

%% file: sections/appendix/evaluation_details.tex
\FloatBarrier
\section{Evaluation Details and Supporting Results}
\label{app:evaluation_details}

\subsection{Evaluation Metrics and Protocols}

For Public Vision, let $s_j$ denote the benchmark's chance-adjusted score on task $j$ and $\omega_j$ its relative weight. The aggregate is
\begin{equation}
S_{\mathrm{Public}}=\frac{\sum_j\omega_j s_j}{\sum_j\omega_j}.
\end{equation}
The nine core tasks have unit weight; R-Bench-M and MMMU-Pro Vision have relative weights 0.5 and 0.25. Full is the equal-weight mean of Public and Private. All official Vision results use the snapshot specified in the opening footnote \ref{fig:public_vision}. Appendix~\ref{app:full_public_rankings} gives the expanded Full rankings; Table~\ref{tab:full_public} summarizes Piccolo's score components.

\begin{table}[!htbp]
\centering\small
\caption{Official Vision score components. Difference is Public minus Full, in points.}
\label{tab:full_public}
\begin{tabularx}{\linewidth}{@{}Yrrrr@{}}
\toprule
\textbf{Model} & \textbf{Full} & \textbf{Private} & \textbf{Public} & \textbf{Difference} \\
\midrule
\textcolor{picink}{\textbf{Piccolo-27B}} & 75.59 & 70.97 & 80.22 & 4.63 \\
\textcolor{picink}{\textbf{Piccolo-9B}} & 68.35 & 65.04 & 71.67 & 3.32 \\
\textcolor{picink}{\textbf{Piccolo-4B}} & 67.46 & 63.49 & 71.42 & 3.96 \\
\bottomrule
\end{tabularx}
\end{table}

\subsection{Application Comparisons}
\label{app:application-series}

\begin{figure}[!htbp]
\centering
\includegraphics[width=\linewidth]{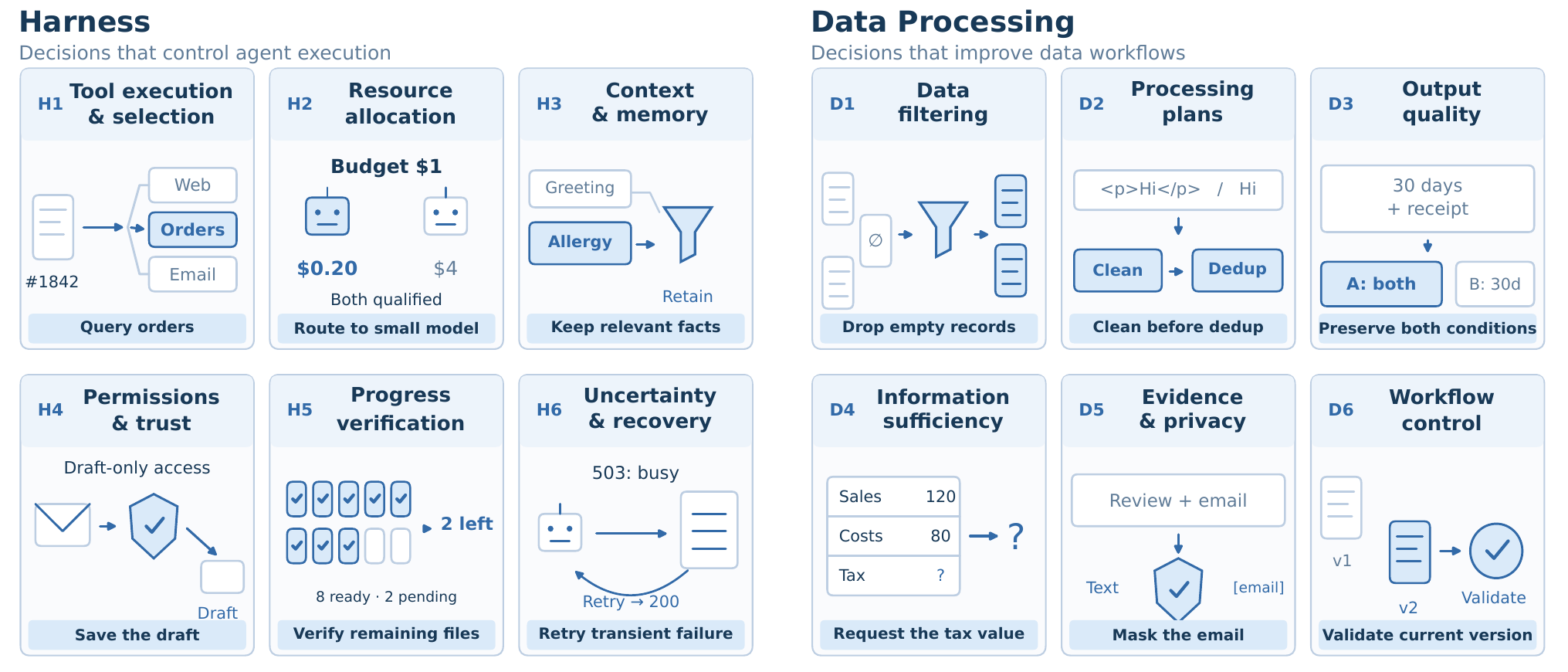}
\caption{Full Harness and Data capability illustrations. The twelve authored scenes show representative decisions about execution and data state, adapted from the project website.}
\label{fig:workflow_details}
\end{figure}

The application evaluations cover the hard-target subset of HarnessJevBench v0.5 (295 tasks, 303 questions), 500 DataJevBench questions, and 231 JevBench questions. Table~\ref{tab:application-full} extends the main comparison to all evaluated baselines. Each external model uses its published input format, classification head, attention mechanism, and temperature, together with the official scoring scripts. Figure~\ref{fig:workflow_details} illustrates the full Harness and Data capability sets.
\begin{table}[!htb]
\centering\small
\caption{Application scores on HarnessJevBench, DataJevBench, and JevBench-231. Bold and underlined values mark the best and second-best scores per column.}
\label{tab:application-full}
\setlength{\tabcolsep}{4pt}
\renewcommand{\arraystretch}{1.08}
\begin{tabularx}{\linewidth}{@{}Yrrr@{}}
\toprule
\textbf{Model} & \textbf{HarnessJevBench} & \textbf{DataJevBench} & \textbf{JevBench} \\
\midrule
\textcolor[HTML]{3C5578}{\textbf{Piccolo-27B}} & \textbf{84.41} & 82.60 & 92.21 \\
JDI-Decision-V1 & 82.71 & \textbf{84.40} & 90.48 \\
WeMM-JEV-27B & 82.03 & 82.40 & 92.21 \\
RBDash-Decision v0.2 & 72.37 & 79.60 & 88.31 \\
Valen-Decision 1.5 27B & 79.15 & 80.20 & 91.34 \\
Perplexity Decider v1.3 (27B) & 82.03 & \underline{83.00} & 90.91 \\
Amos 1 Large (27B) & 80.68 & 81.80 & \underline{93.07} \\
Sieve 31B v3 & \underline{83.05} & 80.00 & 90.48 \\
StartLux-Decision v1.1 27B & 82.03 & 79.60 & \textbf{93.94} \\
Valen-Decision 1.5 9B & 64.58 & 66.40 & 86.15 \\
D3 Flash 9B & 65.42 & 68.20 & 82.25 \\
\textcolor[HTML]{3C5578}{\textbf{Piccolo-9B}} & 71.53 & 72.40 & 88.74 \\
\textcolor[HTML]{3C5578}{\textbf{Piccolo-4B}} & 66.10 & 70.20 & 87.01 \\
Valen-Decision 1.5 4B & 62.03 & 65.40 & 83.55 \\
D3 Mini 4B & 65.93 & 65.80 & 81.82 \\
\textcolor[HTML]{3C5578}{\textbf{Piccolo-2B}} & 56.10 & 52.80 & 86.15 \\
\textcolor[HTML]{3C5578}{\textbf{Piccolo-0.8B}} & 51.19 & 52.20 & 75.32 \\
\bottomrule
\end{tabularx}
\end{table}

\subsection{Public Vision Task Profiles}
\label{app:vision-profiles}

Table~\ref{tab:vision_complete} reports all eleven Public task scores. The first group follows the six leading Full-score entries; the remaining groups compare smaller models. Piccolo-27B exceeds JDI-Decision V1 by 1.78 points on CV-Bench, 4.02 on BLINK, and 3.38 on MMMU-Pro Vision, as highlighted in Section~\ref{sec:overall-results}.

\begin{table}[!htb]
\centering\footnotesize
\caption{Public Vision task scores for leading models and smaller-model peers. Both panels use the official snapshot of Figure~\ref{fig:public_vision}; higher scores are better. Bold and underlined values mark the best and second-best scores per task among the displayed models. KIE combines CORD and FUNSD; Moderation uses Hateful Memes.}
\label{tab:vision_complete}
\setlength{\tabcolsep}{3pt}
\renewcommand{\arraystretch}{1.13}
\begin{tabular*}{\linewidth}{@{\extracolsep{\fill}}lrrrrrr@{}}
\toprule
\textbf{Model} & CV-Bench & BLINK & \shortstack{RealWorld\\QA} & CharXiv & \shortstack{Infographic\\VQA} & Mind2Web \\
\midrule
\textcolor{picink}{\textbf{Piccolo-27B}} & \textcolor{picink}{\textbf{83.41}} & \textcolor{picink}{\underline{82.76}} & \textcolor{picink}{72.53} & \textcolor{picink}{80.95} & \textcolor{picink}{\textbf{98.91}} & \textcolor{picink}{90.96} \\
JDI-Decision V1 & \underline{81.63} & 78.74 & \textbf{74.75} & \underline{81.95} & \underline{98.36} & 88.90 \\
WeMM-JEV-27B & 81.38 & 74.22 & \underline{74.20} & 81.70 & \underline{98.36} & 85.89 \\
RBDash v0.2 & 77.82 & 75.35 & 73.36 & 79.20 & 97.81 & 87.67 \\
Valen 1.5 (27B) & 78.14 & \textbf{85.82} & 70.87 & 80.95 & 97.81 & \textbf{93.29} \\
H2O Lightning 31B v1.3 & 80.82 & 57.79 & 66.98 & \textbf{86.97} & 97.27 & 86.98 \\
\midrule
\textcolor{picink}{\textbf{Piccolo-9B}} & \textcolor{picink}{80.65} & \textcolor{picink}{58.60} & \textcolor{picink}{67.54} & \textcolor{picink}{76.44} & \textcolor{picink}{93.44} & \textcolor{picink}{85.75} \\
Valen 1.5 (9B) & 77.09 & 77.45 & 68.65 & 77.19 & 93.99 & \textbf{93.29} \\
D3 Flash 9B & 75.39 & 56.02 & 62.27 & 73.93 & 96.17 & 80.82 \\
\midrule
\textcolor{picink}{\textbf{Piccolo-4B}} & \textcolor{picink}{81.22} & \textcolor{picink}{66.33} & \textcolor{picink}{62.54} & \textcolor{picink}{72.43} & \textcolor{picink}{92.90} & \textcolor{picink}{83.01} \\
PPLX Flash v1.2 (4B) & 75.64 & 55.54 & 63.10 & 76.94 & 93.99 & \underline{92.87} \\
D3 Mini 4B & 73.94 & 53.44 & 57.83 & 73.93 & 94.54 & 77.53 \\
\bottomrule
\end{tabular*}
\par\vspace{9pt}
\begin{tabular*}{\linewidth}{@{\extracolsep{\fill}}lrrrrr@{}}
\toprule
\textbf{Model} & Winoground & KIE & Moderation & R-Bench-M & \shortstack{MMMU-Pro\\Vision} \\
\midrule
\textcolor{picink}{\textbf{Piccolo-27B}} & \textcolor{picink}{\textbf{88.50}} & \textcolor{picink}{\underline{93.79}} & \textcolor{picink}{53.01} & \textcolor{picink}{47.30} & \textcolor{picink}{\textbf{54.59}} \\
JDI-Decision V1 & 86.75 & 93.66 & 56.80 & 43.15 & \underline{51.21} \\
WeMM-JEV-27B & \underline{87.00} & \underline{93.79} & 52.51 & 38.46 & 46.23 \\
RBDash v0.2 & \underline{87.00} & 92.99 & 57.63 & 35.21 & 44.44 \\
Valen 1.5 (27B) & 86.00 & \textbf{96.59} & \textbf{74.44} & \textbf{67.33} & 42.65 \\
H2O Lightning 31B v1.3 & 86.25 & 91.85 & 58.45 & 41.34 & 51.01 \\
\midrule
\textcolor{picink}{\textbf{Piccolo-9B}} & \textcolor{picink}{83.75} & \textcolor{picink}{92.79} & \textcolor{picink}{33.22} & \textcolor{picink}{31.42} & \textcolor{picink}{43.58} \\
Valen 1.5 (9B) & 82.00 & \textbf{96.59} & \underline{72.30} & \underline{64.99} & 35.69 \\
D3 Flash 9B & 82.75 & 92.92 & 40.97 & 20.77 & 38.28 \\
\midrule
\textcolor{picink}{\textbf{Piccolo-4B}} & \textcolor{picink}{83.50} & \textcolor{picink}{92.65} & \textcolor{picink}{34.87} & \textcolor{picink}{33.58} & \textcolor{picink}{40.46} \\
PPLX Flash v1.2 (4B) & 83.50 & 91.72 & 54.49 & 23.48 & 31.51 \\
D3 Mini 4B & 81.50 & 92.19 & 37.35 & 18.78 & 32.71 \\
\bottomrule
\end{tabular*}
\end{table}

\subsection{Development Feedback and Validation}
\label{app:feedback_validation}

The data-allocation study uses about 4,000 tuning questions and 2,000 disjoint confirmation questions. The tuning set supports repeated allocation decisions; confirmation readouts assess transfer to other questions before broad benchmark evaluation. Source, group, and construction identifiers define split membership and keep related variants together. Fixed-question readouts retain their inputs and targets, while training-window summaries describe the examples encountered during training.

The branches in Figure~\ref{fig:adaptive-data} have comparable sample-exposure budgets. Their recorded trajectories include differences in starting parameter states and concurrent changes to several data allocations. The displayed outcomes characterize those complete research paths. On the 341 fixed workflow questions in Table~\ref{tab:proposal}, target--prediction KL and net correctness changes jointly inform the subsequent mixture update.

\subsection{Selected Comparison Protocols}
\label{app:training-inference}
\label{app:discussion-records}

The joint-training comparison in Figure~\ref{fig:recipe_gains}a holds input and inference adaptation fixed, including the prompt attention mask and image-associated OCR. The backbone comparison in panel (b) uses a shared recipe and data order. The attention comparison in Section~\ref{sec:model-evolution} uses the same Text development questions across mask combinations; the objective comparison starts from a shared model with the same examples and update budget.

For synthesized tasks, quality verification checks the target, while model pass rate measures current difficulty (Table~\ref{tab:synthesis}). Family identifiers keep related constructions together when splitting examples. These measurements guide seed revision and the admission of further synthesized data.

\begin{table}[!htbp]
\centering\small
\caption{Model pass rates on quality-verified synthetic tasks. Pass rate measures the model's current success on each task family.}
\label{tab:synthesis}
\begin{tabularx}{\linewidth}{@{}lYr@{}}
\toprule
\textbf{Task family} & \textbf{Decision focus} & \textbf{Pass rate} \\
\midrule
State tracking & Track changes to a described state & 63.43\% \\
Argument validation & Check tool-call argument constraints & 100.00\% \\
Budget allocation & Allocate limited resources under constraints & 15.38\% \\
\bottomrule
\end{tabularx}
\end{table}

The input-transfer comparisons in Figure~\ref{fig:transfer-exposure}a preserve model weights within each pair. Panel (b) follows eleven checkpoints on a fixed pool through 1.024 million sample exposures, counting repeated presentations. The archive also retains unsuccessful continuations and their parameter differences; selected update components can be reused in later combinations, as discussed in Section~\ref{sec:discussion}.

%% file: sections/appendix/research_records_compute.tex
\FloatBarrier
\section{Compute Accounting}
\label{app:compute_accounting}

The archived 27B campaign comprises 647 training attempts and approximately 6,300 accelerator-hours on Alibaba PPU Zhenwu 810E devices (\texttt{PPU-ZW810E})~\cite{ppu810e}. Accelerator-hours sum occupied devices multiplied by training duration across attempts. This accounting covers the archived 27B training campaign; subsequent multimodal and smaller-model training, evaluation, synthesis, and candidate-search workloads are recorded separately.

To estimate peak-equivalent compute, we use a reference BF16 rate of $p_{\mathrm{ref}}=123$ TFLOP/s per device.\footnote{The reference rate is listed in the ZW810E specification table of a publicly reposted \href{https://www.scribd.com/document/1010774773}{presentation}. It is the throughput assumption for this estimate.} For $H\approx6{,}300$ accelerator-hours,
\begin{equation}
C_{\mathrm{peak}}=H\times3{,}600\times p_{\mathrm{ref}}\times10^{12}
\simeq2.8\times10^{21}\ \text{FLOP},
\label{eq:compute_accounting}
\end{equation}
corresponding to approximately 32 PFLOP/s-days. This estimate expresses accelerator time at the reference peak rate; executed FLOPs depend on achieved throughput. The training volumes in Appendix~\ref{app:data_recipes} describe individual curricula, while this compute total aggregates the exploration attempts.

\paragraph{Training scales in related work.}
Table~\ref{tab:training_scale_context} summarizes reported training scales for decision models. Piccolo's representative multimodal curricula use approximately 380K or 700K examples (Appendix~\ref{app:data_recipes}), supporting iterative refinement with fewer than one million examples per recipe.

\begin{table}[!htbp]
\centering\small
\caption{Reported training scales and resource accounting. Each row retains the scope and units of its source.}
\label{tab:training_scale_context}
\begin{tabularx}{\linewidth}{@{}>{\raggedright\arraybackslash}p{.20\linewidth}>{\raggedright\arraybackslash}p{.27\linewidth}Y@{}}
\toprule
\textbf{Project} & \textbf{Reported scope} & \textbf{Data and resources} \\
\midrule
\textcolor{picink}{\textbf{Piccolo}} & Archived 27B exploration & 647 attempts; approximately 6,300 PPU-ZW810E accelerator-hours. \\
\addlinespace[2pt]
Mammoth-1 / Elephant-1~\cite{mammoth} & Released 27B / 9B SFT recipes & 6.21M examples, one epoch; 72 GB300 / 128 H100 accelerators; 9,859 / 5,546 updates. \\
\addlinespace[2pt]
StartLux-Decision~\cite{startlux} & Full SFT and adapter stages & Stage 1: 0.8M examples for 27B, 2.7M for 0.8B--9B; Stage 2: 1.7M records. \\
\addlinespace[2pt]
Drex Max~\cite{drex} & SFT, low-rank RLAF, vision SFT & 1.15M, 180K, and 95K records; 9,000, 1,200, and 1,500 updates, respectively. \\
\bottomrule
\end{tabularx}
\end{table}

%% file: references.bib
@misc{drex,
      title={Towards Foundation Decision Models: Drex}, 
      author={Nischay Dhankhar and Alisher Amantay and Zhanibek Datbayev and Darkhan Akhmetov and Yash Shah and Nurmukhamed Ubaidullayev and Raman Saparkhan and Dos Baha and Abulhair Saparov},
      year={2026},
      eprint={2611.07854},
      archivePrefix={arXiv},
      primaryClass={cs.CL},
      url={https://arxiv.org/abs/2611.07854}, 
}

@misc{startlux,
      title={StartLux-Decision: AutoThink Decision Models with Reinforcement Learning for Calibrated Deliberative Decisions}, 
      author={Xialie Zhuang and Zunhai Su and Jia Pan and Ngai Wong and Guanhua Chen and Chuan-Wei Kuo},
      year={2026},
      eprint={2610.17309},
      archivePrefix={arXiv},
      primaryClass={cs.CL},
      url={https://arxiv.org/abs/2610.17309}, 
}

@misc{datajuicer,
      title={Data-Juicer: A One-Stop Data Processing System for Large Language Models}, 
      author={Daoyuan Chen and Yilun Huang and Zhijian Ma and Hesen Chen and Xuchen Pan and Ce Ge and Dawei Gao and Yuexiang Xie and Zhaoyang Liu and Jinyang Gao and Yaliang Li and Bolin Ding and Jingren Zhou},
      year={2023},
      eprint={2309.02033},
      archivePrefix={arXiv},
      primaryClass={cs.LG},
      url={https://arxiv.org/abs/2309.02033}, 
}

@inproceedings{ade,
  title     = {Agentic Data Engineering for LLMs: System Design and a Controlled Empirical Study},
  author    = {Ding, Ruomeng and Shen, Qianli and Han, Zhaoyang and Chen, Meixin and Chen, Daoyuan and Li, Yaliang},
  booktitle = {NeurIPS 2026 Workshop on Automated Machine Learning Research (AutoMLR)},
  year      = {2026},
  note      = {arXiv:2610.14238 [cs.CL]},
  url       = {https://arxiv.org/abs/2610.14238}
}

@misc{cues,
      title={CuES: A Curiosity-driven and Environment-grounded Synthesis Framework for Agentic RL}, 
      author={Shinji Mai and Yunpeng Zhai and Ziqian Chen and Cheng Chen and Anni Zou and Shuchang Tao and Zhaoyang Liu and Bolin Ding},
      year={2025},
      eprint={2512.01311},
      archivePrefix={arXiv},
      primaryClass={cs.AI},
      url={https://arxiv.org/abs/2512.01311}, 
}

@misc{dgm,
      title={Darwin Godel Machine: Open-Ended Evolution of Self-Improving Agents}, 
      author={Jenny Zhang and Shengran Hu and Cong Lu and Robert Lange and Jeff Clune},
      year={2026},
      eprint={2505.22954},
      archivePrefix={arXiv},
      primaryClass={cs.AI},
      url={https://arxiv.org/abs/2505.22954}, 
}

@misc{evolutionarymerge,
      title={Evolutionary Optimization of Model Merging Recipes}, 
      author={Takuya Akiba and Makoto Shing and Yujin Tang and Qi Sun and David Ha},
      year={2025},
      eprint={2403.13187},
      archivePrefix={arXiv},
      primaryClass={cs.NE},
      doi={https://doi.org/10.1038/s42256-024-00975-8},
      url={https://arxiv.org/abs/2403.13187}, 
}

@inproceedings{sandbox,
  title     = {Data-Juicer Sandbox: A Feedback-Driven Suite for Multimodal Data-Model Co-development},
  author    = {Chen, Daoyuan and Wang, Haibin and Huang, Yilun and Ge, Ce and Li, Yaliang and Ding, Bolin and Zhou, Jingren},
  booktitle = {Proceedings of the 42nd International Conference on Machine Learning (ICML)},
  year      = {2025}
}

@misc{harnessbench,
  author={{HarnessJevBench Contributors}},
  title={{HarnessJevBench: Benchmarking Jev-style Models at Agent Harness Decision Points}},
  year={2026},
  url={https://github.com/agentscope-ai/HarnessJevBench}
}

@misc{databench,
  author={{DataJevBench Contributors}},
  title={{DataJevBench}},
  year={2026},
  url={https://github.com/datajuicer/DataJevBench}
}

@misc{piccoloweb,
  author={{Piccolo Team}},
  title={{Piccolo: Decision Models for Harnesses, Data Workflows, and Games}},
  year={2026},
  url={https://decisionplayground.agentscope.io/piccolo.html}
}

@misc{mammoth,
 title={{Mammoth-1 and Elephant-1: frontier decision models}},
 author={Max Kirkby and Michael Psenka and Thomas Villeneuve and Aaryam Sharma and Charles O'Neill},
 year={2026}, url={https://labs.baseten.co/articles/mammoth-1-and-elephant-1-frontier-decision-models}}

@article{multiclasscalibration,
  author  = {Ambuj Tewari and Peter L. Bartlett},
  title   = {On the Consistency of Multiclass Classification Methods},
  journal = {Journal of Machine Learning Research},
  year    = {2007},
  volume  = {8},
  number  = {36},
  pages   = {1007--1025},
  url     = {http://jmlr.org/papers/v8/tewari07a.html}
}

@article{multiclasslogloss,
  title = {Statistical Analysis of Some Multi-Category Large Margin Classification Methods},
  author = {Zhang, Tong},
  year = {2004},
  journal = {Journal of Machine Learning Research},
  volume = {5},
  pages = {1225--1251},
  url = {https://jmlr.org/papers/v5/zhang04b.html},
}

@misc{decisionboard,
  author={{Decision Index Contributors}},
  title={{Decision Index: Vision and Text Leaderboards}},
  year={2026},
  url={https://huggingface.co/spaces/multimodalart/jev-decision-index}
}

@misc{agentscope,
      title={AgentScope 1.0: A Developer-Centric Framework for Building Agentic Applications},
      author={Dawei Gao and Zitao Li and Yuexiang Xie and Weirui Kuang and Liuyi Yao and Bingchen Qian and Zhijian Ma and Yue Cui and Haohao Luo and Shen Li and Lu Yi and Yi Yu and Shiqi He and Zhiling Luo and Wenmeng Zhou and Zhicheng Zhang and Xuguang He and Ziqian Chen and Weikai Liao and Farruh Isakulovich Kushnazarov and Yaliang Li and Bolin Ding and Jingren Zhou},
      year={2025},
      eprint={2508.16279},
      archivePrefix={arXiv},
      primaryClass={cs.AI},
      url={https://arxiv.org/abs/2508.16279},
}

@misc{qwenpawpiccolo,
  author={{AgentScope Team}},
  title={{QwenPaw: Local Piccolo Decisions}},
  year={2026},
  url={https://github.com/agentscope-ai/QwenPaw/blob/main/docs/local-piccolo.md}
}

@misc{ppu810e,
  author={{Alibaba Cloud}},
  title={{PPU: Zhenwu 810E Product Specifications}},
  url={https://help.aliyun.com/zh/cs/user-guide/ppu/},
  year={2026},
  note={Accessed November 13, 2026}
}

@misc{mopd,
      title={MOPD: Multi-Teacher On-Policy Distillation for Capability Integration in LLM Post-Training},
      author={Wenhan Ma and Jianyu Wei and Liang Zhao and Hailin Zhang and Bangjun Xiao and Lei Li and Qibin Yang and Bofei Gao and Yudong Wang and Rang Li and Jinhao Dong and Zhifang Sui and Fuli Luo},
      year={2026},
      eprint={2606.30406},
      archivePrefix={arXiv},
      primaryClass={cs.CL},
      url={https://arxiv.org/abs/2606.30406},
}

@inproceedings{data_banking77,
  author = {I{\~{n}}igo Casanueva and Tadas Temcinas and Daniela Gerz and Matthew Henderson and Ivan Vulic},
  title = {{Efficient Intent Detection with Dual Sentence Encoders}},
  booktitle = {Proceedings of the 2nd Workshop on NLP for ConvAI - ACL 2020},
  year = {2020},
  url = {https://arxiv.org/abs/2003.04807},
}

@inproceedings{data_clinc150,
  author = {Larson, Stefan and Mahendran, Anish and Peper, Joseph J. and Clarke, Christopher and Lee, Andrew and Hill, Parker and Kummerfeld, Jonathan K. and Leach, Kevin and Laurenzano, Michael A. and Tang, Lingjia and Mars, Jason},
  title = {{An Evaluation Dataset for Intent Classification and Out-of-Scope Prediction}},
  booktitle = {Proceedings of the 2019 Conference on Empirical Methods in Natural Language Processing and the 9th International Joint Conference on Natural Language Processing (EMNLP-IJCNLP)},
  year = {2019},
  url = {https://www.aclweb.org/anthology/D19-1131},
}

@misc{data_esci,
  author = {Chandan K. Reddy and Lluís Màrquez and Fran Valero and Nikhil Rao and Hugo Zaragoza and Sambaran Bandyopadhyay and Arnab Biswas and Anlu Xing and Karthik Subbian},
  title = {{Shopping Queries Dataset: A Large-Scale {ESCI} Benchmark for Improving Product Search}},
  year = {2022},
  eprint = {2206.06588},
  archivePrefix = {arXiv},
  url = {https://arxiv.org/abs/2206.06588},
}

@inproceedings{data_go_emotions,
  author = {Demszky, Dorottya and Movshovitz-Attias, Dana and Ko, Jeongwoo and Cowen, Alan and Nemade, Gaurav and Ravi, Sujith},
  title = {{GoEmotions: A Dataset of Fine-Grained Emotions}},
  booktitle = {58th Annual Meeting of the Association for Computational Linguistics (ACL)},
  year = {2020},
  url = {https://huggingface.co/datasets/google-research-datasets/go_emotions},
}

@inproceedings{data_isarcasmeval,
  author = {Abu Farha, Ibrahim and Oprea, Silviu Vlad and Wilson, Steven and Magdy, Walid},
  title = {{SemEval-2022 Task 6: iSarcasmEval, Intended Sarcasm Detection in English and Arabic}},
  booktitle = {Proceedings of the 16th International Workshop on Semantic Evaluation (SemEval-2022)},
  year = {2022},
  pages = {802--814},
  doi = {10.18653/v1/2022.semeval-1.111},
  url = {https://aclanthology.org/2022.semeval-1.111/}
}

@misc{data_massive,
  author = {Jack FitzGerald and Christopher Hench and Charith Peris and Scott Mackie and Kay Rottmann and Ana Sanchez and Aaron Nash and Liam Urbach and Vishesh Kakarala and Richa Singh and Swetha Ranganath and Laurie Crist and Misha Britan and Wouter Leeuwis and Gokhan Tur and Prem Natarajan},
  title = {{MASSIVE: A 1M-Example Multilingual Natural Language Understanding Dataset with 51 Typologically-Diverse Languages}},
  year = {2022},
  eprint = {2204.08582},
  archivePrefix = {arXiv},
  url = {https://arxiv.org/abs/2204.08582},
}

@inproceedings{data_mnli,
  author = {Williams, Adina and Nangia, Nikita and Bowman, Samuel},
  title = {{A Broad-Coverage Challenge Corpus for Sentence Understanding through Inference}},
  booktitle = {Proceedings of the 2018 Conference of the North American Chapter of the Association for Computational Linguistics: Human Language Technologies, Volume 1 (Long Papers)},
  year = {2018},
  pages = {1112--1122},
  publisher = {Association for Computational Linguistics},
  url = {http://aclweb.org/anthology/N18-1101},
}

@inproceedings{data_sgd,
  author = {Rastogi, Abhinav and Zang, Xiaoxue and Sunkara, Srinivas and Gupta, Raghav and Khaitan, Pranav},
  title = {{Towards scalable multi-domain conversational agents: The schema-guided dialogue dataset}},
  booktitle = {Proceedings of the AAAI Conference on Artificial Intelligence},
  year = {2020},
  volume = {34},
  pages = {8689--8696},
  url = {https://github.com/google-research-datasets/dstc8-schema-guided-dialogue},
}

@inproceedings{data_snli,
  author = {Bowman, Samuel R. and Angeli, Gabor and Potts, Christopher and Manning, Christopher D.},
  title = {{A large annotated corpus for learning natural language inference}},
  booktitle = {Proceedings of the 2015 Conference on Empirical Methods in Natural Language Processing},
  year = {2015},
  pages = {632--642},
  publisher = {Association for Computational Linguistics},
  doi = {10.18653/v1/D15-1075},
  url = {https://aclanthology.org/D15-1075},
}

@misc{data_wanli,
  author = {Liu, Alisa and Swayamdipta, Swabha and Smith, Noah A. and Choi, Yejin},
  title = {{WANLI: Worker and AI Collaboration for Natural Language Inference Dataset Creation}},
  year = {2022},
  url = {https://arxiv.org/pdf/2201.05955},
}

@inproceedings{data_aqua_rat,
  author = {Ling, Wang and Yogatama, Dani and Dyer, Chris and Blunsom, Phil},
  title = {{Program induction by rationale generation: Learning to solve and explain algebraic word problems}},
  booktitle = {ACL},
  year = {2017},
  url = {https://huggingface.co/datasets/deepmind/aqua_rat},
}

@article{data_arc,
  author = {Peter Clark and Isaac Cowhey and Oren Etzioni and Tushar Khot and Ashish Sabharwal and Carissa Schoenick and Oyvind Tafjord},
  title = {{Think you have Solved Question Answering? Try ARC, the AI2 Reasoning Challenge}},
  journal = {arXiv:1803.05457v1},
  year = {2018},
  url = {https://huggingface.co/datasets/allenai/ai2_arc},
}

@inproceedings{data_boolq,
  author = {Clark, Christopher and Lee, Kenton and Chang, Ming-Wei and Kwiatkowski, Tom and Collins, Michael and Toutanova, Kristina},
  title = {{BoolQ: Exploring the Surprising Difficulty of Natural Yes/No Questions}},
  booktitle = {NAACL},
  year = {2019},
  url = {https://huggingface.co/datasets/google/boolq},
}

@inproceedings{data_commonsenseqa,
  author = {Talmor, Alon and Herzig, Jonathan and Lourie, Nicholas and Berant, Jonathan},
  title = {{CommonsenseQA: A Question Answering Challenge Targeting Commonsense Knowledge}},
  booktitle = {Proceedings of the 2019 Conference of the North {A}merican Chapter of the Association for Computational Linguistics: Human Language Technologies, Volume 1 (Long and Short Papers)},
  year = {2019},
  pages = {4149--4158},
  publisher = {Association for Computational Linguistics},
  doi = {10.18653/v1/N19-1421},
  eprint = {1811.00937},
  archivePrefix = {arXiv},
  url = {https://aclanthology.org/N19-1421},
}

@inproceedings{data_cosmos_qa,
  author = {Huang, Lifu and Le Bras, Ronan and Bhagavatula, Chandra and Choi, Yejin},
  title = {{Cosmos {QA}: Machine Reading Comprehension with Contextual Commonsense Reasoning}},
  booktitle = {Proceedings of the 2019 Conference on Empirical Methods in Natural Language Processing and the 9th International Joint Conference on Natural Language Processing (EMNLP-IJCNLP)},
  year = {2019},
  pages = {2391--2401},
  publisher = {Association for Computational Linguistics},
  doi = {10.18653/v1/D19-1243},
  url = {https://www.aclweb.org/anthology/D19-1243},
}

@inproceedings{data_finqa,
  author = {Chen, Zhiyu and Chen, Wenhu and Smiley, Charese and Shah, Sameena and Borova, Iana and Langdon, Dylan and Moussa, Reema and Beane, Matt and Huang, Ting-Hao and Routledge, Bryan and Wang, William Yang},
  title = {{FinQA: A Dataset of Numerical Reasoning over Financial Data}},
  booktitle = {Proceedings of EMNLP 2021},
  year = {2021},
  url = {https://github.com/czyssrs/FinQA},
}

@inproceedings{data_hellaswag,
  author = {Zellers, Rowan and Holtzman, Ari and Bisk, Yonatan and Farhadi, Ali and Choi, Yejin},
  title = {{HellaSwag: Can a Machine Really Finish Your Sentence?}},
  booktitle = {Proceedings of the 57th Annual Meeting of the Association for Computational Linguistics},
  year = {2019},
  url = {https://huggingface.co/datasets/Rowan/hellaswag},
}

@article{data_musique_train,
  author = {Trivedi, Harsh and Balasubramanian, Niranjan and Khot, Tushar and Sabharwal, Ashish},
  title = {{MuSiQue: Multihop Questions via Single-hop Question Composition}},
  journal = {Transactions of the Association for Computational Linguistics},
  year = {2022},
  volume = {10},
  pages = {539--554},
  doi = {10.1162/tacl_a_00475},
  url = {https://aclanthology.org/2022.tacl-1.31/}
}

@article{data_qasc,
  author = {Tushar Khot and Peter Clark and Michal Guerquin and Peter Jansen and Ashish Sabharwal},
  title = {{QASC: A Dataset for Question Answering via Sentence Composition}},
  journal = {arXiv:1910.11473v2},
  year = {2020},
  url = {https://huggingface.co/datasets/allenai/qasc},
}

@article{data_winogrande,
  author = {Sakaguchi, Keisuke and Le Bras, Ronan and Bhagavatula, Chandra and Choi, Yejin},
  title = {{WinoGrande: An Adversarial Winograd Schema Challenge at Scale}},
  journal = {arXiv preprint arXiv:1907.10641},
  year = {2019},
  url = {https://arxiv.org/abs/1907.10641}
}

@misc{data_bekko_selected,
  author = {{hotchpotch}},
  title = {{Bekko System One Dataset v0}},
  year = {2026},
  howpublished = {Hugging Face dataset repository},
  url = {https://huggingface.co/datasets/hotchpotch/bekko-system-one-dataset-v0},
  note = {Accessed November 18, 2026},
}

@misc{data_helpsteer2,
  author = {Zhilin Wang and Yi Dong and Olivier Delalleau and Jiaqi Zeng and Gerald Shen and Daniel Egert and Jimmy J. Zhang and Makesh Narsimhan Sreedhar and Oleksii Kuchaiev},
  title = {{HelpSteer2: Open-source dataset for training top-performing reward models}},
  year = {2024},
  eprint = {2406.08673},
  archivePrefix = {arXiv},
  url = {https://arxiv.org/abs/2406.08673},
}

@misc{data_prm800k,
  author = {Hunter Lightman and Vineet Kosaraju and Yura Burda and Harri Edwards and Bowen Baker and Teddy Lee and Jan Leike and John Schulman and Ilya Sutskever and Karl Cobbe},
  title = {{Let's Verify Step by Step}},
  year = {2023},
  eprint = {2305.20050},
  archivePrefix = {arXiv},
  url = {https://arxiv.org/abs/2305.20050}
}

@misc{data_prometheus,
  author = {Seungone Kim and Jamin Shin and Yejin Cho and Joel Jang and Shayne Longpre and Hwaran Lee and Sangdoo Yun and Seongjin Shin and Sungdong Kim and James Thorne and Minjoon Seo},
  title = {{Prometheus: Inducing Fine-grained Evaluation Capability in Language Models}},
  year = {2023},
  eprint = {2310.08491},
  archivePrefix = {arXiv},
  url = {https://arxiv.org/abs/2310.08491},
}

@misc{data_ragtruth,
  author = {Cheng Niu and Yuanhao Wu and Juno Zhu and Siliang Xu and Kashun Shum and Randy Zhong and Juntong Song and Tong Zhang},
  title = {{RAGTruth: A Hallucination Corpus for Developing Trustworthy Retrieval-Augmented Language Models}},
  year = {2024},
  eprint = {2401.00396},
  archivePrefix = {arXiv},
  url = {https://arxiv.org/abs/2401.00396v2},
}

@misc{data_jevforge_mind2web,
  author = {{AndeyTait}},
  title = {{JevForge-Mind2Web}},
  year = {2026},
  howpublished = {Hugging Face dataset repository},
  url = {https://huggingface.co/datasets/AndeyTait/JevForge-Mind2Web},
  note = {Accessed November 18, 2026},
}

@misc{data_procedural_typed_decisions,
  author = {{tasksource}},
  title = {{Procedural Typed Decisions}},
  year = {2026},
  howpublished = {Hugging Face dataset repository},
  url = {https://huggingface.co/datasets/tasksource/procedural-typed-decisions},
  note = {Accessed November 18, 2026},
}

@misc{data_when2call,
  author = {{NVIDIA}},
  title = {{When2Call}},
  year = {2025},
  howpublished = {Hugging Face dataset repository},
  url = {https://huggingface.co/datasets/nvidia/When2Call},
  note = {Accessed November 18, 2026},
}

@misc{data_eikos_decisions,
  author = {Caio Vicentino},
  title = {{Eikos Decisions: training data for open, calibrated, single-pass typed-decision models}},
  year = {2026},
  howpublished = {Hugging Face dataset repository},
  url = {https://huggingface.co/datasets/caiovicentino1/eikos-decisions},
  note = {Accessed November 18, 2026},
}

@misc{data_helmo_synthetic,
  author = {{helmo}},
  title = {{Synthetic Typed Decisions (Helmo)}},
  year = {2026},
  howpublished = {Hugging Face dataset repository},
  url = {https://huggingface.co/datasets/helmo/synthetic-typed-decisions},
  note = {Accessed November 18, 2026},
}

@misc{data_jev_decisions_clean50k,
  author = {{samatv256}},
  title = {{Jev Decisions v1}},
  year = {2026},
  howpublished = {Hugging Face dataset repository},
  url = {https://huggingface.co/datasets/samatv256/jev-decisions-v1},
  note = {Accessed November 18, 2026},
}

@misc{data_jev_distill_v3,
  author = {{SargeDev}},
  title = {{Jev Distill Corpus v3}},
  year = {2026},
  howpublished = {Hugging Face dataset repository},
  url = {https://huggingface.co/datasets/SargeDev/jev-distill-corpus-v3},
  note = {Accessed November 18, 2026},
}

@misc{data_minojev_data,
  author = {{zeredy879}},
  title = {{MinoJev Data}},
  year = {2026},
  howpublished = {Hugging Face dataset repository},
  url = {https://huggingface.co/datasets/zeredy879/minojev-data},
  note = {Accessed November 18, 2026},
}

@misc{data_open_jev,
  author = {{ZefanCai}},
  title = {{Open-Jev}},
  year = {2026},
  howpublished = {Hugging Face dataset repository},
  url = {https://huggingface.co/datasets/ZefanCai/Open-Jev},
  note = {Accessed November 18, 2026},
}

@misc{data_open_jev_v11,
  author = {{ZefanCai}},
  title = {{Open-Jev v1.1}},
  year = {2026},
  howpublished = {Hugging Face dataset repository},
  url = {https://huggingface.co/datasets/ZefanCai/Open-Jev-v1.1},
  note = {Accessed November 18, 2026},
}

@misc{data_plumb_decisions,
  author = {{crh225}},
  title = {{Plumb Decisions}},
  year = {2026},
  howpublished = {Hugging Face dataset repository},
  url = {https://huggingface.co/datasets/crh225/plumb-decisions},
  note = {Accessed November 18, 2026},
}

@misc{data_synthetic_typed_decisions,
  author = {{tasksource}},
  title = {{Synthetic Typed Decisions (tasksource)}},
  year = {2026},
  howpublished = {Hugging Face dataset repository},
  url = {https://huggingface.co/datasets/tasksource/synthetic-typed-decisions},
  note = {Accessed November 18, 2026},
}

@misc{data_tasksource_jev,
  author = {{tasksource}},
  title = {{Tasksource JEV Typed Decisions}},
  year = {2026},
  howpublished = {Hugging Face dataset repository},
  url = {https://huggingface.co/datasets/tasksource/tasksource-jev-typed-decisions},
  note = {Accessed November 18, 2026},
}

@misc{data_typed_decisions,
  author = {{LocalLLaMA}},
  title = {{Typed Decisions}},
  year = {2026},
  howpublished = {Hugging Face dataset repository},
  url = {https://huggingface.co/datasets/LocalLLaMA/typed-decisions},
  note = {Accessed November 18, 2026},
}

@misc{data_typed_decisions_ja,
  author = {{hiroki-abe-58}},
  title = {{Typed Decisions (Japanese): typed-decisions-ja v3}},
  year = {2026},
  howpublished = {Hugging Face dataset repository},
  url = {https://huggingface.co/datasets/GeneLab/typed-decisions-ja},
  note = {Accessed November 18, 2026},
}

@misc{data_typed_decisions_pt_es,
  author = {{telepatia-ai}},
  title = {{Typed Decisions (Portuguese and Spanish)}},
  year = {2026},
  howpublished = {Hugging Face dataset repository},
  url = {https://huggingface.co/datasets/telepatia-ai/typed-decisions-pt-es},
  note = {Accessed November 18, 2026},
}

@misc{data_typed_decisions_synth,
  author = {Muhammed Nazeem},
  title = {{Typed Decisions Synth}},
  year = {2026},
  howpublished = {Hugging Face dataset repository},
  url = {https://huggingface.co/datasets/n4ze3m/typed-decisions-synth},
  note = {Accessed November 18, 2026},
}

@misc{data_mind2web_original,
  author = {Xiang Deng and Yu Gu and Boyuan Zheng and Shijie Chen and Samuel Stevens and Boshi Wang and Huan Sun and Yu Su},
  title = {{Mind2Web: Towards a Generalist Agent for the Web}},
  year = {2023},
  eprint = {2306.06070},
  archivePrefix = {arXiv},
  url = {https://arxiv.org/abs/2306.06070}
}

@inproceedings{data_tasksource_original,
  author = {Sileo, Damien},
  title = {{tasksource: A Large Collection of {NLP} tasks with a Structured Dataset Preprocessing Framework}},
  booktitle = {Proceedings of the 2024 Joint International Conference on Computational Linguistics, Language Resources and Evaluation (LREC-COLING 2024)},
  year = {2024},
  pages = {15655--15684},
  publisher = {ELRA and ICCL},
  doi = {10.63317/2h4pkr685mkq},
  url = {https://aclanthology.org/2024.lrec-main.1361},
}

@article{data_reasoning_core,
  author = {Sileo, Damien and Lacombe, Valentin and Kachler, Dimitri},
  title = {{Reasoning Core: Designing Broad Procedural Data for Completion-Supervised Reasoning Training}},
  journal = {arXiv preprint arXiv:2608.05148},
  year = {2026},
  url = {https://arxiv.org/abs/2608.05148},
}
